%% file: main.tex
\pdfoutput=1
\documentclass[11pt]{article}

\usepackage[final]{acl}

\usepackage{times}
\usepackage{latexsym}

\usepackage[T1]{fontenc}
\usepackage[utf8]{inputenc}

\usepackage{microtype}

\usepackage{inconsolata}

\usepackage{graphicx}
\usepackage{xcolor}   % For additional colors

\usepackage{hyperref}
\usepackage{url}
\usepackage{algorithm}
\usepackage{algorithmic}
\usepackage{multirow}
\usepackage{wrapfig}
\usepackage{makecell}
\usepackage{tikz}
\usetikzlibrary{ducks}
\usepackage{subcaption}
\usepackage{booktabs}
\usepackage{hhline}
\usepackage{amssymb}
\usepackage{enumitem}
\input{./math_commands}

\title{Towards Robust and Efficient Federated Low-Rank Adaptation \\
with Heterogeneous Clients}

\newif\ifsubmit
\submitfalse % Default: not submitted

\newcommand{\submit}[1]{%
    \ifnum#1=1%
        \global\submittrue%
    \else%
        \global\submitfalse%
    \fi%
}

\submit{1}

\author{
 \textbf{Jabin Koo\thanks{Equal Contribution}\textsuperscript{1}},
 \textbf{Minwoo Jang\footnotemark[1]\textsuperscript{2}},
 \textbf{Jungseul Ok\thanks{Corresponding Author}\textsuperscript{1, 2}}
\\
\\
 \textsuperscript{1}Department of Computer Science and Engineering, POSTECH, South Korea \\
 \textsuperscript{2}Graduate School of Artificial Intelligence, POSTECH, South Korea
\\
 \small{
   \{\href{mailto:jbkoo@postech.ac.kr}{jbkoo}, \href{mailto:jbkoo@postech.ac.kr}{minwoo}, \href{mailto:jbkoo@postech.ac.kr}{jungseul}\}@postech.ac.kr
 }
}

\begin{document}

\maketitle
\begin{abstract}
\label{sec:abs}
\input{./0_abstract}
\end{abstract}

\section{Introduction}
\label{sec:intro}

\input{./1_intro}

\section{Related Works}
\label{sec:related}
\input{./2_related_works}

\section{Problem Formulation}
\label{sec:problem}
\input{./3_problem}

\section{Proposed Method}
\label{sec:methodology}
\input{./4_methodology}

\section{Experiments}
\label{sec:experiments}
\input{./5_experiments}

\section{Conclusion}
\label{sec:conclusion}
\input{./6_conclusion}

\section{Acknowledgments}
\label{sec:acknowledgments}
\input{./acknowledgments}

\section{Limitations}
\label{sec:limitations}
\input{./7_limitation}

\bibliography{main}

\clearpage
\appendix

\section{Dataset Statistics}
\label{sec:appendix_a}
\input{./appendix_a}

\section{Reproducibility}
\label{sec:appendix_b}
\input{./appendix_b}

\section{Additional Experiments}
\label{sec:appendix_c}
\input{./appendix_c}

\section{Theoretical Proofs}
\label{sec:appendix_d}
\input{./appendix_d}

\end{document}

%% file: math_commands.tex
\usepackage{amsmath,amsfonts,bm}

\def\1{\bm{1}}

\DeclareMathAlphabet{\mathsfit}{\encodingdefault}{\sfdefault}{m}{sl}
\SetMathAlphabet{\mathsfit}{bold}{\encodingdefault}{\sfdefault}{bx}{n}

%% file: 0_abstract.tex
Federated fine-tuning for Large Language Models (LLMs) faces significant challenges due to the heavy communication overhead of transmitting large model updates. Although Low Rank Adaptation (LoRA) has been proposed as a solution, yet its application in federated learning is complicated by discordance in aggregation. Existing methods addressing this discordance often suffer from performance degradation at low ranks in heterogeneous data settings. In response, we introduce LoRA-A$^2$ (Low Rank Adaptation with Alternating freeze and Adaptive rank selection), which demonstrates robustness in challenging settings with low ranks and high data heterogeneity. Our experimental findings reveal that LoRA-A$^2$ maintains performance even under extreme heterogeneity and low rank conditions, achieving up to a significant reduction in uploaded parameters compared to full fine-tuning without compromising performance. This adaptive mechanism increases robustness and communication efficiency in federated fine-tuning, enabling the practical deployment of LLMs in resource-constrained environments.

%% file: 1_intro.tex
\begin{figure*}[t]
    \centering
    \includegraphics[width=0.95\textwidth]{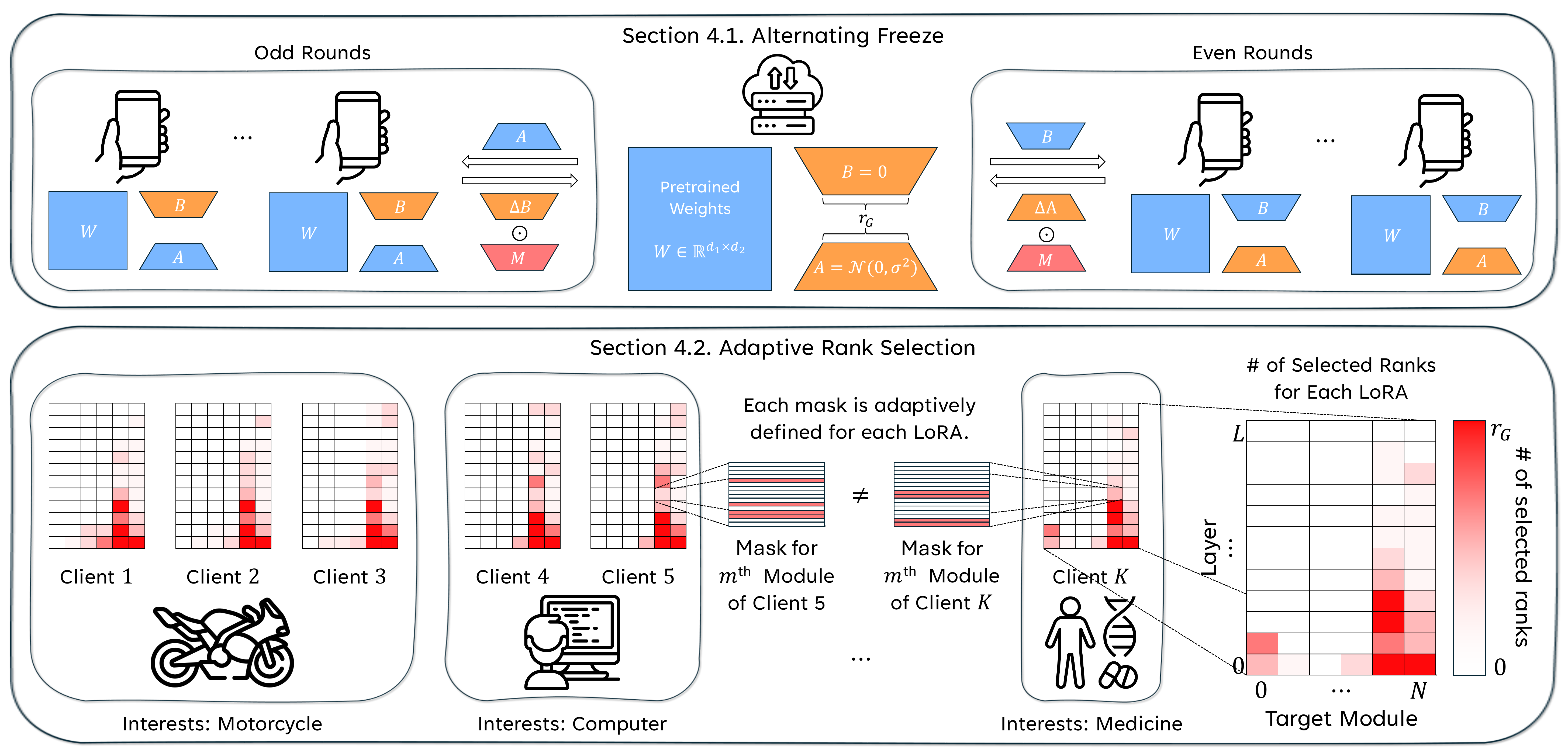} 
    \caption{An overview of the proposed method, LoRA-A$^2$. It alternately trains $B$ and $A$ of the LoRA adapters, with each client training only a subset of the downloaded parameters. LoRA-A$^2$ is free from several issues for using LoRA in FL, which are discussed in Section \ref{sec:problem}. A detailed explanation of the method is provided in Section \ref{sec:methodology}.}
    \label{fig:method_fig}
\end{figure*}

Large Language Models (LLMs), exemplified by ChatGPT \citep{openai2024gpt4technicalreport}, Llama \citep{dubey2024llama} and others, represent a hallmark of the current era. These models are being widely applied in real-world scenarios by fine-tuning them on various task-specific datasets \cite{dodge2020fine}. With the expansion of edge devices, the potential to leverage rich, privacy-sensitive data for fine-tuning LLMs has shifted the focus toward federated fine-tuning. Despite its potential, this is often infeasible due to the large size of LLMs, which require extensive computational and communication resources from local devices.

Parameter-Efficient Fine-Tuning (PEFT) methods \cite{lester-etal-2021-power, liu2022few} are increasingly being explored in the context of federated fine-tuning. Among these, Low-Rank Adaptation (LoRA) \citep{hu2022lora} is particularly noteworthy for its significant reduction in number of communicated parameters. However, naive application of LoRA in Federated Learning (FL) \citep{pmlr-v54-mcmahan17a} environment comes with several challenges such as aggregation discordance. Although several solutions have been proposed, they often remain vulnerable to high heterogeneity and low ranks due to a limited parameter space, making it difficult to reduce rank size for communication efficiency in realistic FL scenarios.

To address this, we introduce LoRA-A$^2$ (\textbf{Lo}w \textbf{R}ank \textbf{A}daptation with \textbf{A}lternating freeze and \textbf{A}daptive rank selection), which is robust to both high heterogeneity and low ranks. LoRA-A$^2$ incorporates two main strategies: (1) \textbf{alternating freeze}, which switches between freezing LoRA modules $B$ and $A$ in each round, and (2) \textbf{adaptive rank selection}, which identifies and updates only important ranks in LoRA modules.
We conduct experiments across various rank sizes and heterogeneity levels, comparing our algorithm with multiple baselines. Through the experiments, we reveal the vulnerabilities of existing methods and highlight the robustness of LoRA-A$^2$ in challenging conditions, providing analyses of the reasons for its robustness. Additionally, we empirically demonstrate that our approach achieves performance comparable to or exceeding that of full fine-tuning, while uploading less than 0.2\% of parameters to the server.
\vspace{1em}

\noindent Our contributions can be summarized as follows:
\begin{itemize} [leftmargin=10pt]
    \item We address the vulnerabilities of previous federated LoRA methods in high heterogeneity and low-rank settings, and propose a novel algorithm, LoRA-A$^2$, which demonstrates robustness in these challenging conditions.
    \item Our algorithm effectively reduces communication costs, achieving a significant reduction in uploaded parameters compared to full fine-tuning, while maintaining its performance.
    \item We provide visualization on adaptive rank selection process and a thorough empirical exploration on how important ranks are efficiently trained and transmitted.
\end{itemize}

%% file: 2_related_works.tex
\paragraph{LoRA with adaptive rank selection}
LoRA \citep{hu2022lora} is a widely used PEFT method for LLMs. It tries to approximate the updated part of the pre-trained model with two smaller matrices. This approach is inspired by previous studies \citep{li2018measuring, aghajanyan-etal-2021-intrinsic}, which suggest that newly learned parameters for adaptation lie within a low dimensional subspace.

AdaLoRA \citep{zhang2023adaptive} assumes a scenario where the total parameter budget is limited. It adaptively selects the rank for each LoRA adapter under this constraint, with a criterion for rank selection based on singular values of the updated part.

ALoRA \citep{liu-etal-2024-alora} utilizes a router for each LoRA adapter. The router determines which part of each LoRA adapter should be either turned on or off, enabling efficient fine-tuning via pruning. Similarly, DoRA \cite{mao-etal-2024-dora} re-splits LoRA into smaller groups of LoRAs. During the training session, it estimates the importance of each small LoRA, allowing the parts with less contribution across the whole LoRA to be pruned. Our research extends these adaptive rank selection methods in centralized learning to the FL setting so that each client adaptively selects different ranks suitable for their own dataset.

\paragraph{Federated learning with LoRA}

As training LLMs on mobile devices becomes feasible, fine-tuning LLMs via FL has recently gained attention. In line with this trend, using LoRA for federated fine-tuning \citep{babakniya2023slora, kuo2024federated, wang2024flora}, is also being considered. However, simply adopting LoRA for FL presents several obstacles, which are discussed in Section \ref{sec:problem}.

HetLoRA \citep{cho2023heterogeneous} assumes that each client may have different computational power, which is a common scenario in FL. Based on this assumption, it allows each client to use a LoRA adapter with a different rank. Zero-padding is then applied to align the dimensions of the client-specific adapters for aggregation. 

\citet{sun2024improving} point out that aggregating the two matrices of a LoRA adapter separately cannot fully approximate the original LoRA adapter. Based on this finding, they propose FFA-LoRA, which addresses this issue by freezing half of each LoRA throughout the entire fine-tuning session.

FlexLoRA \citep{bai2024federated} aggregates the product of two matrices comprising each LoRA adapter and then decomposes the aggregated parameters back into two smaller matrices via Singular Value Decomposition (SVD). This approach allows FlexLoRA to overcome the challenges addressed by HetLoRA and FFA-LoRA, respectively, though at the expense of increased computational cost on the server-side for the decomposition process.

%% file: 3_problem.tex
\paragraph{Low rank adaptation}

Because LLMs have billions of parameters, fine-tuning them for specific domains demands significant computational power, which may be infeasible in many situations. To address this issue, PEFT techniques such as LoRA \cite{hu2022lora} have recently gained attention, as they can effectively reduce the number of parameters that need to be trained. Specifically, when fine-tuning a pre-trained weight matrix $W_0 \in \mathbb{R}^{d_1 \times d_2}$ to obtain $W$, LoRA achieves this by decomposing $\Delta W$, the update of the weight matrix, into smaller matrices $B \in \mathbb{R}^{d_1 \times r}$ and $A \in \mathbb{R}^{r \times d_2}$:
\begin{equation}
    \label{eq:lora}
    W = W_{0} + \Delta W = W_{0} + BA,
\end{equation}
where $r \ll \{ d_1, d_2 \}$ denotes the rank of LoRA. With this approximation, the number of trainable parameters is reduced from $d_1 \cdot d_2$ to $r \cdot (d_1 + d_2)$.

\paragraph{Federated LoRA and discordance problem}
Consider a global pre-trained model $W_{0}$ and a set of clients $\{1, 2, \cdots, K\}$. The objective in federated fine-tuning is to update $W_{0}$ to obtain a model $W$ that is suitable for all local datasets $\{\mathcal{D}_k\}_{k=1}^K$. 
However, fine-tuning LLMs is very expensive for local devices in terms of both computation and communication, as billions of parameters must be trained and transmitted in each round.

LoRA presents a promising approach in FL for reducing communication costs, as only low rank matrices $B$ and $A$ are trained and transmitted, allowing the number of communicated parameters to be linearly reduced by the rank $r$ of LoRA modules. However, the straightforward application of LoRA in FL introduces a significant issue known as discordance \citep{sun2024improving}, primarily due to aggregation algorithms. In methods like FedAvg \citep{pmlr-v54-mcmahan17a}, where each weight is aggregated individually, discordance occurs between the actual and aggregated parameters. That is,
\begin{equation}
    \begin{aligned} \label{eq:discordance}
    \sum_{k=1}^K w_k \Delta W_k & = \sum_{k=1}^K w_k B_k A_k \\ & \neq \left( \sum_{k=1}^K w_k B_k \right) \left( \sum_{k=1}^K w_k A_k \right)
\end{aligned}
\end{equation}
in general, where $\sum_{k=1}^K w_k = 1$ with $w_k \geq 0$ for all $k \in [K]$. One might consider aggregating $\Delta W_k = B_kA_k$ directly to eliminate the discordance, but this approach involves decomposing $\Delta W = \sum_{k=1}^K w_k \Delta W_k$ back into $B$ and $A$ for the next round, which is computationally unstable. 
\begin{figure}[t]
    \centering
    \begin{minipage}[b]{0.23\textwidth}
        \centering
        \includegraphics[width=\textwidth]{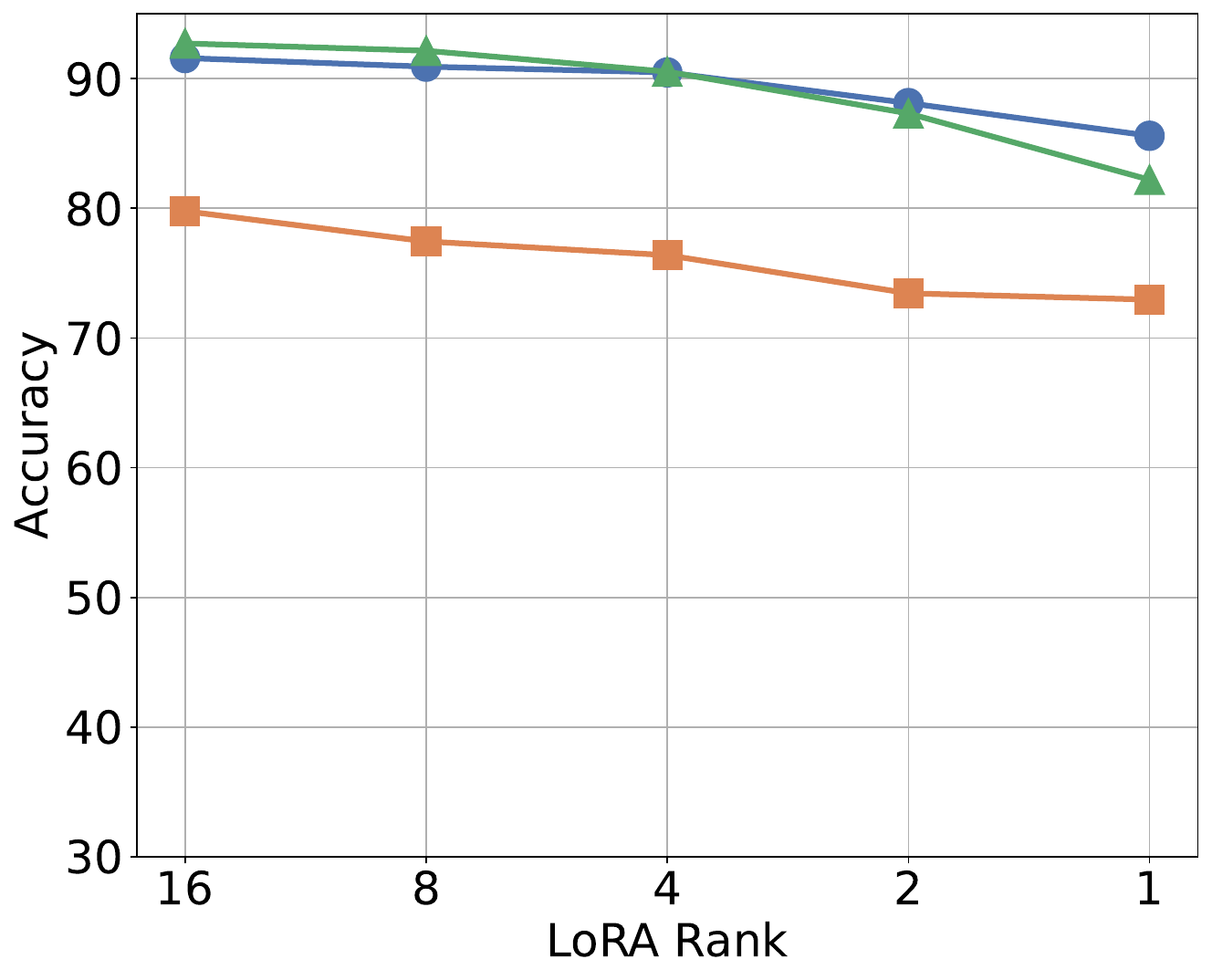} 
        \subcaption{$Dir(0.1)$}
    \end{minipage}
    \hfill
    \begin{minipage}[b]{0.23\textwidth}
        \centering
        \includegraphics[width=\textwidth]{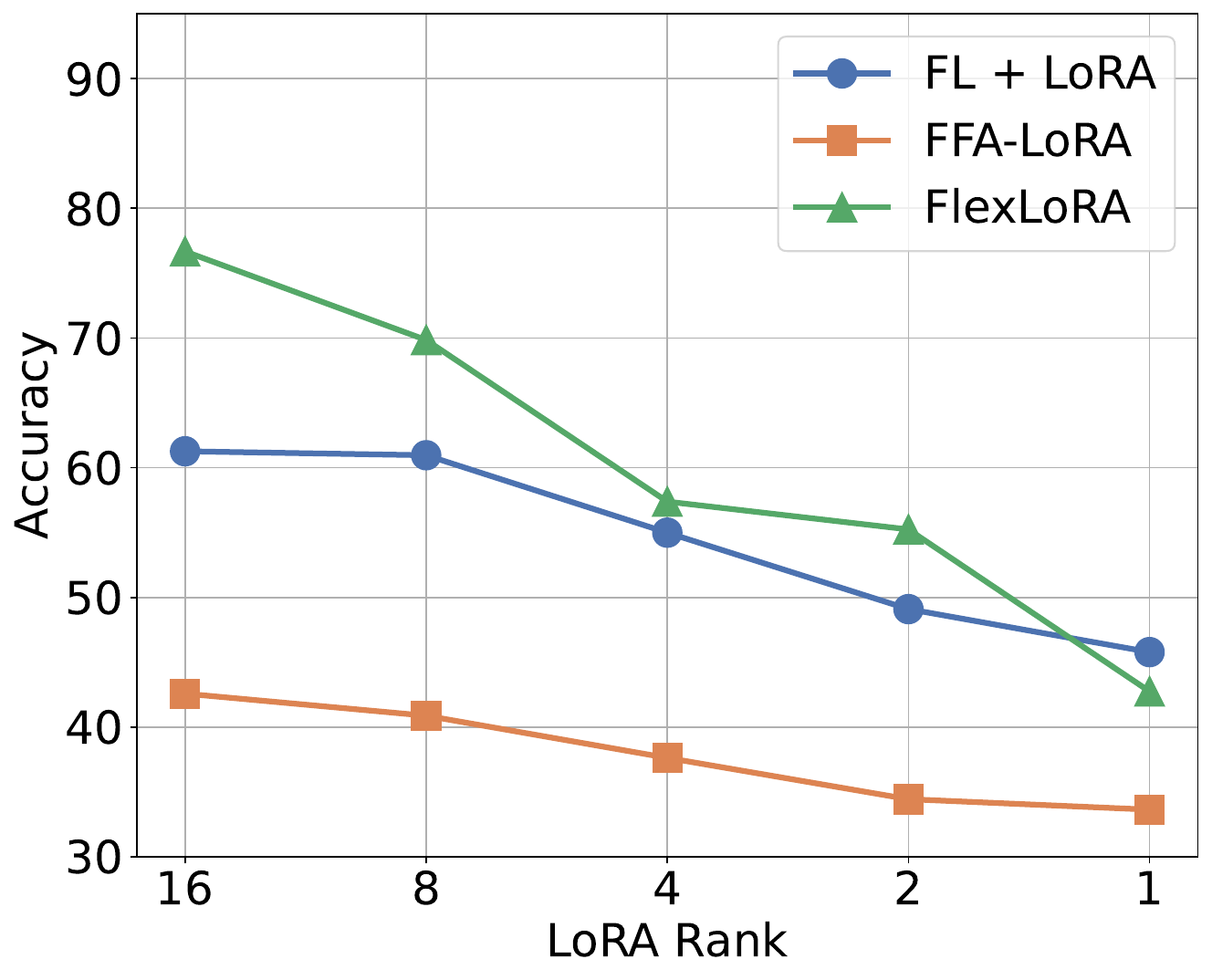} 
        \subcaption{$Dir(0.01)$}
    \end{minipage}
    \caption{Accuracy of previous Federated LoRA methods across different rank sizes in heterogeneous data settings.}
    \label{fig:volnerability}
\end{figure}

\paragraph{Limited parameter space in low rank and high data heterogeneity}
This discrepancy can be effectively addressed by either freezing the LoRA module A, as suggested by \citet{sun2024improving}, or employing SVD decomposition, as outlined by \citet{bai2024federated}. However, Figure \ref{fig:volnerability} illustrates that the accuracy of these approaches decreases significantly at lower ranks under high heterogeneity. We attribute this decline primarily to the restricted parameter space imposed by LoRA. A limited training parameter space constrains the optimization capabilities for complex FL tasks, and a restricted aggregation parameter space exacerbates conflicts among clients. A detailed analysis of this limited parameter space is provided in Appendix \ref{sec:appendix_c}.

%% file: 4_methodology.tex
To tackle the identified challenges, we propose a novel framework called \textbf{Lo}w \textbf{R}ank \textbf{A}daptation with \textbf{A}lternating freeze and \textbf{A}daptive rank selection for federated learning, or LoRA-A$^2$, for communication efficient FL with LoRA. LoRA-A$^2$ adaptively selects LoRA ranks for local training and transmits only the selected part of each adapter in an alternating way.

\subsection{Alternating Freeze}

LoRA-A$^2$ efficiently addresses the issue of discordance by employing a simple alternating freeze technique to train the LoRA modules $B$ and $A$. Instead of solely training module $B$ while keeping module $A$ frozen permanently, as suggested by FFA-LoRA \citep{sun2024improving}, LoRA-A$^2$ alternates between the two: LoRA module $A$ is frozen during even rounds, while module $B$ is frozen during odd rounds. This method preserves the optimization space while effectively resolving discordance. Specifically, when freezing $A$, we have
\begin{equation}
\begin{aligned}
    \Delta W &= \sum_{k=1}^K \left( w_k B_k \right) A \\&= \sum_{k=1}^K \left( w_k B_k A_k \right) = \sum_{k=1}^K \left( w_k \Delta W_k \right),
\end{aligned}
\end{equation}
and the same applies when freezing $B$. In this way, LoRA-A$^2$ trains both $B$ and $A$, ensuring that $A$ does not remain the same as its initial value.

To further enhance the effect of alternating optimization, we adopt different learning rates for $B$ and $A$, inspired by LoRA+ \citep{hayou2024loraplus}. Figure \ref{fig:ffa_alter} demonstrates the effectiveness of alternating freeze and learning rate adjustment.
% that alternatively optimizing LoRA modules is effective than letting A fixed permanently.

\subsection{Adaptive Rank Selection} \label{ars}
Furthermore, we propose an adaptive rank selection method designed to reduce the number of transmitted parameters while preserving the training and aggregation parameter space. This approach selects important LoRA ranks to match local communication rank budget $r_i$ out of global LoRA adapter with rank $r_G$ adaptively based on the local dataset of each clients. We mainly focus on communication cost for uploading parameters to the server as it is known that upload bandwidth is generally much slower than download bandwidth and is the major part of communication cost \cite{konevcny2016federated, suresh2017distributed, kairouz2021advances}.

The adaptive rank selection process provides two key benefits: (1) it minimizes client conflicts by  allowing each client to select different LoRA ranks in high heterogeneity, and (2) it reallocates rank resources from less important LoRA modules to modules that require more fine-tuning, which is especially effective when the communication rank budget is small.

To quantify which ranks are more important, we introduce our original criterion $S_{m,i}$ for each rank $i$ within module $m$ as follows: 

\begin{equation}
\begin{aligned}
    S_{m,i}^{B_k} = \| \Delta {B_k}_{[:,i]} A_{[i,:]} \| _F \\
    S_{m,i}^{A_k} = \| B_{[:,i]} \Delta {A_k}_{[i,:]} \| _F
\end{aligned} \; .
\label{eq:importance_score}
\end{equation}

We define the change in $\Delta W$ for each rank $i$ and module $m$ as contribution $(C_{m,i})$, represented as $\sum C_{m,i}=\Delta W_k^{t+1} - \Delta W_k^{t} =\sum(\Delta {B_k}_{[:,i]} A_{[i,:]})$. And define our criterion $S_{m,i}$ as the Frobenius norm of contribution $(C_{m,i})$. This criterion captures the impact of each rank on model updates based on local gradients. This approach is better suited for LoRA modules than simpler gradient magnitude-based criteria, $||\Delta {B_k}_{[:,i]}||$ or $||\Delta {A_k}_{[i,:]}||$, as our criterion explicitly accounts for the interplay between module A and B. The ablation study in Table \ref{tab:criteria_table} empirically supports the superiority of this criterion. At each round, participating clients run local training for 1 epoch to obtain $\Delta W$ for calculating the contribution.

After computing $S_{m,i}^{B_k}$ or $S_{m,i}^{A_k}$ for each module $m$, we select $\text{top-}(r_i \cdot N)$ LoRA ranks from a total of $r_G \cdot N$ based on the scores across the entire model, where $N$ denotes the number of target modules across all the layers of the base model. We refer to the set of selected ranks of client $k$ as $\mathcal{R}_k$.

Once the ranks are selected, each client defines LoRA module mask $M_k^{(m)}$ for the module $m$ to be
\begin{equation}
\begin{aligned}
    {M_k}^{(m)}_{[:, i]} &= \begin{cases} 
      \textbf{1}^T_{d_1} & \text{if $i \in \mathcal{R}_k$} \\
      \textbf{0}^T_{d_1} & \text{otherwise}
   \end{cases} \; , \\
   {M_k}^{(m)}_{[i, :]} &= \begin{cases} 
      \textbf{1}_{d_2} & \text{if $i \in \mathcal{R}_k$} \\
      \textbf{0}_{d_2} & \text{otherwise}
   \end{cases} \; ,
\end{aligned}
\label{eq:masking}
\end{equation}
which is producted element-wise to the updated part of $B_k$ (or $A_k$). That is, before each backpropagation step, LoRA-A$^2$ calculates
\begin{equation}
\begin{aligned}
    \Delta {B_k}^{(m)} &\leftarrow \Delta {B_k}^{(m)} \odot {M_k}^{(m)} \\
    \Delta {A_k}^{(m)} &\leftarrow \Delta {A_k}^{(m)} \odot {M_k}^{(m)}
\end{aligned}   
\end{equation}
for each $B_k$ (or $A_k$), where the notation $\odot$ stands for the Hadamard product. After each local training, each client uploads $B_k \odot M_k$ (or $A_k \odot M_k)$, resulting in sparsification and reducing the number of uploaded parameters. Then, the server aggregates the uploaded ones, which are again added to the $B_k$ (or $A_k$) saved two rounds before. Algorithm \ref{alg:ars_server} and \ref{alg:ars_client} provides the detailed pseudocode of our LoRA-A$^2$ algorithm.

\subsection{Theoretical Insights}\label{subsec:theory}

In this section, we provide a brief theoretical analysis of the parameter spaces associated with previous methods and our proposed LoRA-A$^2$ framework. To substantiate our approach, we introduce the following proposition:
\\

\textbf{Proposition 1.} For a model $W$, consider LoRA-based FL algorithms which update $r$ rank parameters per round. Let $\Omega_{\mathcal{A}}$ denote the space of all possible parameter values that an algorithm in $\mathcal{A} \in \left\{ \text{FFA-LoRA}, \text{FL+LoRA}, \text{FlexLoRA}, \text{LoRA-A}^2 \right\}$ can make. Then, we have the following:
\[ \Omega_\text{FFA-LoRA} \subsetneq  \Omega_\text{FL + LoRA} = \Omega_\text{FlexLoRA} \subset  \Omega_{\text{LoRA-A}^2}. \]

The proof of the proposition is provided in Appendix \ref{sec:appendix_d}.

Our algorithm is designed to adaptively select the relevant training and aggregation parameter spaces while concurrently reducing the number of parameters that are updated.

\input{./4_1_alg}

%% file: 4_1_alg.tex
\begin{algorithm}[t]
\begin{algorithmic}[l]
   \caption{LoRA-A$^2$}\label{alg:ars_server}
   \STATE  Initialize $\Delta W = BA$ with $B \in \mathbb{R}^{d_1 \times r_G}$ and $A \in \mathbb{R}^{r_G \times d_2}$ for each LoRA adapter
   \FOR{$t=1, 2, \cdots, T$}
   \STATE Sample participants $\mathcal{K}^{(t)} \subseteq [K]$ for round $t$
   \STATE $w_k = |\mathcal{D}_k| / \left (\sum_{k=1}^K |\mathcal{D}_k| \right)$
   \IF{$t \; \% \; 2 = 1$}
   \FOR{$k=1, 2, \cdots, K$ in parallel}
   \STATE $\Delta B_k^{(t+1)} = \text{LocalTraining}(B^{(t)}, t)$
   \STATE $B^{(t+1)} = B^{(t)} + \sum_{k=1}^K w_k \Delta B_k^{(t+1)}$
   \STATE $A^{(t+1)} = A^{(t)}$
   \ENDFOR
   \ELSE
   \FOR{$k=1, 2, \cdots, K$ in parallel}
   \STATE $\Delta A_k^{(t+1)} = \text{LocalTraining} (A^{(t)}, t)$
   \STATE $A^{(t+1)} = A^{(t)} + \sum_{k=1}^K w_k \Delta A_k^{(t+1)}$
   \STATE $B^{(t+1)} = B^{(t)}$
   \ENDFOR
   \ENDIF

   \ENDFOR
\end{algorithmic}
\end{algorithm}

\begin{algorithm}[t]
\begin{algorithmic}[l]
   \caption{LocalTraining}\label{alg:ars_client}

   \STATE {\bfseries [Rank Selection]}
   \STATE Calculate importance scores following \eqref{eq:importance_score}
   \STATE Define the mask $M_k$ following \eqref{eq:masking}
   \STATE {\bfseries [Local Training]}
   \IF{$t \; \% \; 2 = 1$}
   \STATE $B_k^{(t; \; e-1)} = B^{(t)}$
   \FOR{$e = 1, 2. \cdots, E$}
   \STATE $\Delta B_k^{(t; \; e)} = B_k^{(t; \; e)} - B_k^{(t; \; e-1)}$
   \STATE $\Delta B_k^{(t; \; e)} = \Delta B_k^{(t; \;e)} \odot M_k$
   \STATE Backpropagate $\Delta B_k^{(t; \; e)}$
   \ENDFOR
   \STATE {\bfseries Return: $B_k^{(t; \; E)} - B^{(t)}$}
   \ELSE
   \STATE $A_k^{(t; \; e-1)} = A^{(t)}$
   \FOR{$e = 1, 2. \cdots, E$}
   \STATE $\Delta A_k^{(t; \; e)} = A_k^{(t; \; e)} - A_k^{(t; \; e-1)}$
   \STATE $\Delta A_k^{(t; \; e)} = \Delta A_k^{(t; \;e)} \odot M_k$
   \STATE Backpropagate $\Delta A_k^{(t; \; e)}$
   \ENDFOR
   \STATE {\bfseries Return: $A_k^{(t; \; E)} - A^{(t)}$}
   \ENDIF

\end{algorithmic}
\end{algorithm}

%% file: 5_experiments.tex
\begin{table*}[t]
\centering
\resizebox{\textwidth}{!}{%
\begin{tabular}{c|ccc|ccc|c}
\toprule  \midrule
\multirow{2}{*}{Method} & \multicolumn{3}{c|}{BANKING77 Dataset}                                                                                      & \multicolumn{3}{c|}{20 Newsgroups Dataset}                                                                                      & \multirow{2}{*}{\begin{tabular}[c]{@{}c@{}}Communicated\\ $\text{Parameters}^{*}$\end{tabular}} \\
                        & $Dir(0.5)$ & $Dir(0.1)$ & $Dir(0.01)$        & $Dir(0.5)$ & $Dir(0.1)$ & $Dir(0.01)$        &                                                                                                    \\ \midrule
FL (w/o LoRA)            & 92.76$_{\pm 0.30}$      & 90.29$_{\pm 0.73}$      & 67.58$_{\pm 0.44}$              & 70.93$_{\pm 1.04}$      & 68.82$_{\pm 0.69}$      & 64.41$_{\pm 0.30}$               & 186B                                                                                            \\ \midrule \midrule
FL + LoRA$_{(\text{Rank}=8)}$       & 92.80$_{\pm 0.24}$ & 90.47$_{\pm 0.53}$ & 60.96$_{\pm 1.47}$ & \underline{70.44}$_{\pm 0.28}$ & \underline{67.33}$_{\pm 0.18}$ & 43.90$_{\pm 1.08}$ & 1.99B                                                                                            \\
FFA-LoRA$_{(\text{Rank}=8)}$        & 87.20$_{\pm 0.57}$ & 77.44$_{\pm 1.28}$ & 40.88$_{\pm1.04}$  & 67.00$_{\pm 0.67}$ & 61.27$_{\pm 0.71}$ & 37.34$_{\pm 0.30}$ & 0.991B                                                                                           \\
FlexLoRA$_{(\text{Rank}=8)}$        & \textbf{93.35}$_{\pm 0.24}$ & \textbf{92.14}$_{\pm 0.25}$ & \underline{69.84}$_{\pm 0.65}$ & \textbf{70.59}$_{\pm 0.22}$ & \textbf{68.10}$_{\pm 0.38}$ & \textbf{60.41}$_{\pm 1.54}$                   & 1.99B                                                                                             \\ \midrule
Ours$_{(\text{Rank}=8)}$            & \underline{93.24}$_{\pm 0.27}$                    & \underline{91.61}$_{\pm 0.39}$                     & \textbf{70.13}$_{\pm 1.22}$ & 70.26$_{\pm 0.21}$ & 67.12$_{\pm 0.22}$ & \underline{54.50}$_{\pm 1.44}$ & {1.31B}                                                                                                    \\ \midrule \midrule
FL + LoRA$_{(\text{Rank}=4)}$       & \underline{92.86}$_{\pm 0.08}$ & 88.11$_{\pm 0.88}$ & 54.99$_{\pm 0.59}$ & 70.33$_{\pm 0.12}$ & \underline{67.29}$_{\pm 0.19}$ & 43.12$_{\pm 2.67}$ & 0.991B                                                                                           \\
FFA-LoRA$_{(\text{Rank}=4)}$        & 86.90$_{\pm 1.14}$ & 76.38$_{\pm 0.61}$ & 37.63$_{\pm 0.80}$ & 67.75$_{\pm 0.45}$ & 61.25$_{\pm 0.26}$ & 36.04$_{\pm 0.80}$ & 0.497B                                                                                           \\
FlexLoRA$_{(\text{Rank}=4)}$        & 92.71$_{\pm 0.31}$ & \underline{90.53}$_{\pm 0.70}$ & \underline{57.38}$_{\pm 1.30}$ & 70.05$_{\pm 0.14}$ & \textbf{68.00}$_{\pm 0.33}$ & \underline{50.50}$_{\pm 2.09}$ & 0.991B                                                                                             \\ \midrule 
Ours$_{(\text{Rank}=4)}$            & \textbf{93.22}$_{\pm 0.24}$ & \textbf{91.43}$_{\pm 0.63}$ & \textbf{69.63}$_{\pm 1.52}$ & \textbf{70.28}$_{\pm 0.32}$ & 67.12$_{\pm 0.60}$ & \textbf{53.04}$_{\pm 1.68}$ & {0.888B}                                                                                                \\ \midrule \midrule
FL + LoRA$_{(\text{Rank}=2)}$       & 91.97$_{\pm 0.43}$ & 85.59$_{\pm 1.13}$ & 49.08$_{\pm 0.56}$ & \textbf{70.14}$_{\pm 0.13}$ & 65.40$_{\pm 0.31}$ & 39.07$_{\pm 2.23}$ & 0.497B                                                                                            \\
FFA-LoRA$_{(\text{Rank}=2)}$        & 84.65$_{\pm 1.05}$ & 73.44$_{\pm 0.88}$ & 34.44$_{\pm 2.15}$ & 68.12$_{\pm 0.47}$ & 61.57$_{\pm 0.38}$ & 36.65$_{\pm 0.52}$ & 0.249B                                                                                            \\ 
FlexLoRA$_{(\text{Rank}=2)}$        & \underline{92.22}$_{\pm 0.50}$ & 87.31$_{\pm 0.27}$ & \underline{55.24}$_{\pm 2.19}$ & 70.03$_{\pm 0.31}$ & 66.17$_{\pm 1.70}$ & 48.23$_{\pm 1.73}$ & 0.497B                                                                                             \\ \midrule 
Ours$_{(\text{Rank}=2)}$            & \textbf{93.10}$_{\pm 0.07}$ & \textbf{92.02}$_{\pm 0.36}$                    & \textbf{69.40}$_{\pm 0.48}$ & \underline{70.12}$_{\pm 0.18}$ & \textbf{67.02}$_{\pm 0.26}$ & \textbf{52.99}$_{\pm 2.56}$ & {0.528B}                                                                                              \\ \midrule \midrule
FL + LoRA$_{(\text{Rank}=1)}$       & \underline{90.61}$_{\pm 0.10}$ & \underline{82.24}$_{\pm 1.68}$ & \underline{45.78}$_{\pm 1.04}$ & 69.40$_{\pm 0.33}$ & \underline{63.16}$_{\pm 0.53}$ & \underline{36.58}$_{\pm 0.98}$ & 0.249B                                                                                            \\
FFA-LoRA$_{(\text{Rank}=1)}$        & 82.51$_{\pm 0.53}$ & 72.96$_{\pm 0.54}$ & 33.68$_{\pm 0.20}$ & 67.73$_{\pm 0.30}$ & 61.35$_{\pm 0.22}$ & 34.44$_{\pm 0.68}$ & 0.124B                                                                                            \\ 
FlexLoRA$_{(\text{Rank}=1)}$        & 90.40$_{\pm 0.54}$ & 82.20$_{\pm 0.74}$ & 42.75$_{\pm 0.89}$ & \underline{69.53}$_{\pm 0.25}$ & 62.98$_{\pm 1.12}$ & 35.54$_{\pm 0.68}$ & 0.249B                                                                                             \\ \midrule 
Ours$_{(\text{Rank}=1)}$            & \textbf{93.21}$_{\pm 0.13}$ &        \textbf{91.87}$_{\pm 0.33}$             & \textbf{68.88}$_{\pm 1.15}$ & \textbf{70.31}$_{\pm 0.24}$ & \textbf{66.95}$_{\pm 0.07}$ & \textbf{54.84}$_{\pm 1.15}$ & {0.270B}                                                                                              \\ 
 \midrule \bottomrule
\end{tabular}
}
\caption{Results with RoBERTa-base on BANKING77 and 20 Newsgroups datasets. Smaller $\alpha$ for $Dir(\alpha)$ implies that the simulated setting is more heterogeneous. The best results on each dataset are shown in \textbf{bold} and second best is shown by \underline{underline}. ${}^{*}$ This column reports the total number of uploaded parameters, averaged across rows.}
\label{tab:merged_table}
\end{table*}

In this section, we evaluate the performance of our algorithm against existing FL methods combined with LoRA across various heterogeneity settings and datasets. We assess performance based on accuracy and the total number of uploaded parameters.

\subsection{Experimental Settings}
We mainly adopt pre-trained RoBERTa-base \citep{liu2019roberta} as the base model for fine-tuning. The base model has approximately 125M parameters, all of which are frozen during the fine-tuning phase. And a frozen classifier is added upon the model, following \citet{sun2024improving}. For Table \ref{tab:roberta-large} and \ref{tab:distilbert}, we adopt RoBERTa-large and DistilBERT\citep{sanh2019distilbert}, respectively. RoBERTa-large has approximately 355M parameters, and DistilBERT has approximately 82M parameters.
For fine-tuning, we choose BANKING77 \citep{casanueva-etal-2020-efficient} and 20 Newsgroups \citep{Lang95} datasets.
These datasets are chosen for their ability to simulate a controlled level of data heterogeneity using Dirichlet distribution \citep{hsu2019measuring}. Dataset statistics are reported in Appendix \ref{sec:appendix_a}.

Unless otherwise stated, we trained 30 local clients under full participation, i.e., $\mathcal{K}^{(t)} = [K]$ for all $t \in [T]$. The clients were trained for 50 rounds with 5 local epochs. Detailed hyperparameters for experiments are specified in Appendix \ref{sec:appendix_b}.

For baselines, we adopt four methods that utilize LoRA for federated fine-tuning: FL + LoRA, FFA-LoRA \citep{sun2024improving}, FlexLoRA \citep{bai2024federated}, and HetLoRA \citep{cho2023heterogeneous}, where FL + LoRA stands for the naive implementation of LoRA in FedAvg \citep{pmlr-v54-mcmahan17a}.

\subsection{Main Results} \label{subsec:main_results}
We compare our algorithm with the baseline methods under various data heterogeneity settings in BANKING77 and 20 Newsgroups datasets to demonstrate that our algorithm outperforms previous federated LoRA fine-tuning methods across different non-IID settings and LoRA ranks.

\paragraph{Robustness of LoRA-A$^2$ in low ranks and high heterogeneity}
Table \ref{tab:merged_table} highlights the vulnerability of previous methods under conditions of high heterogeneity and low ranks. The accuracy of baseline methods declines significantly as rank decreases, whereas our algorithm maintains its performance, achieving up to a 23\% accuracy advantage. This suggests that reducing LoRA ranks is challenging for previous methods under realistic heterogeneous data conditions. Also, our algorithm consistently achieves the highest performance or remains within a 1\% margin of the best-performing baselines at rank 8 and 4, while showing a large performance gap in low ranks.
\paragraph{Communication cost reduction by LoRA-A$^2$}
Decreasing LoRA ranks in federated LoRA methods reduces the communication cost linearly. Our algorithm achieves performance comparable to or better than fully fine-tuned models even at rank 1, allowing for up to a 99.8\% reduction in communicated parameters with minimal performance degradation. This demonstrates that LoRA-A$^2$ effectively solves the significant communication cost challenges of federated fine-tuning on LLMs.

\begin{figure*}[htbp]
    \centering
    \begin{minipage}[b]{0.3\textwidth}
        \centering
        \includegraphics[width=\textwidth]{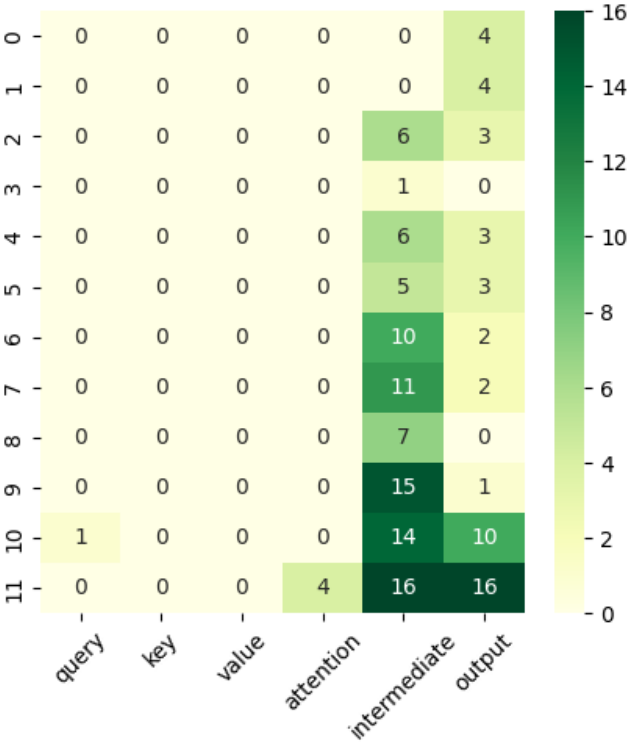}
        \subcaption{client 0}
    \end{minipage}
    \hfill
    \begin{minipage}[b]{0.3\textwidth}
        \centering
        \includegraphics[width=\textwidth]{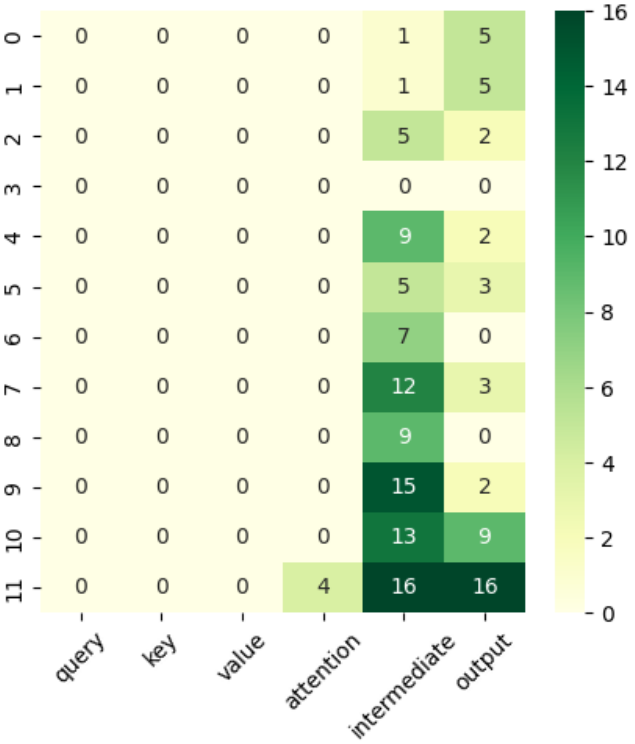}
        \subcaption{client 1}
    \end{minipage}
    \hfill
    \begin{minipage}[b]{0.3\textwidth}
        \centering
        \includegraphics[width=\textwidth]{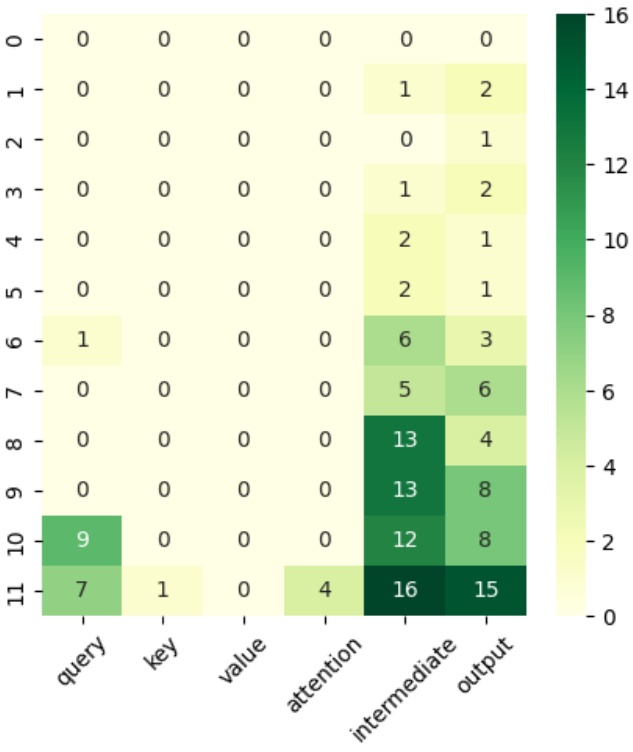}
        \subcaption{client 2}
    \end{minipage}
    \caption{Visualization on number of selected rank per module. The x-axis shows RoBERTa module types, while the y-axis indicates layer numbers. Experimented on the 20 Newsgroups dataset with a pathological data distribution. Average 2 ranks were selected out of 16 ranks by our adaptive rank selection algorithm.}
    \label{fig:module selection}
\end{figure*}

\begin{figure}[t]
    \centering
    \begin{minipage}[b]{0.23\textwidth}
        \centering
        \includegraphics[width=\textwidth]{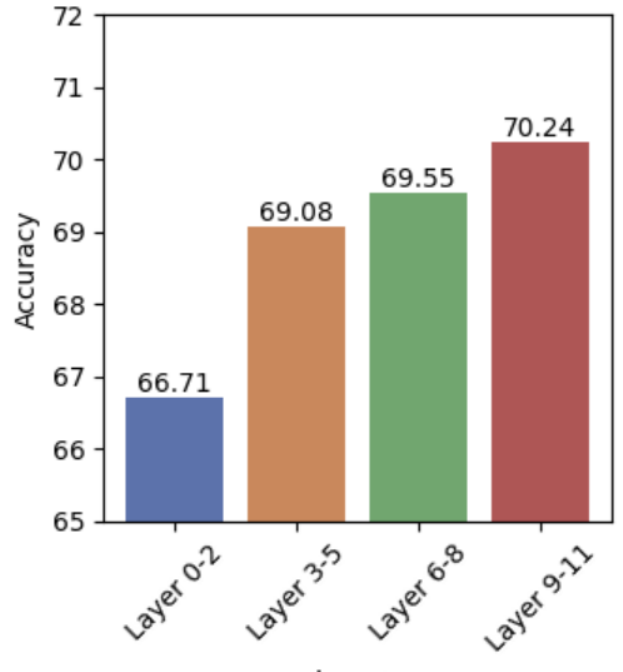} 
        \subcaption{Selected layers}
    \end{minipage}
    \hfill
    \begin{minipage}[b]{0.238\textwidth}
        \centering
        \includegraphics[width=\textwidth]{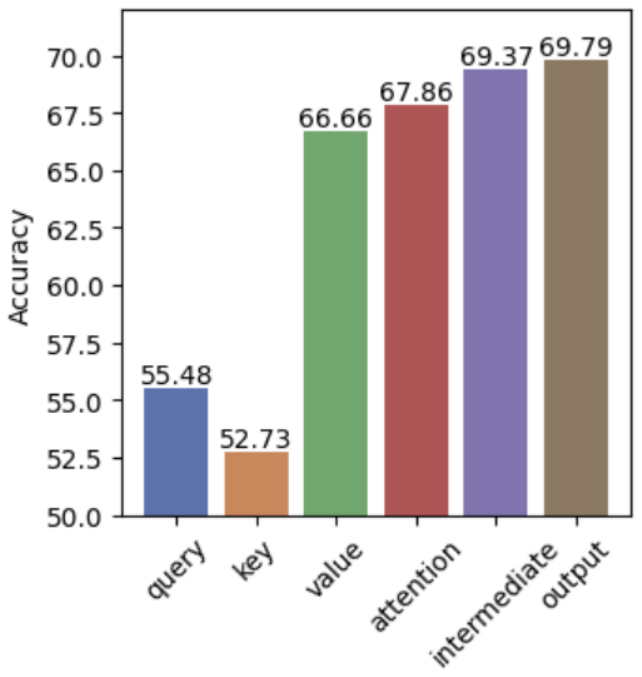} 
        \subcaption{Selected modules}
    \end{minipage}
    \caption{Ablation analysis on the performance of model when solely fine-tuned on selected layers or types of modules. Experimented on 20 Newsgroups dataset with Dir(0.1) heterogeneity.}
    \label{fig:module ablation}
\end{figure}

\begin{figure}[t]
    \centering
    \begin{minipage}[b]{0.39\textwidth}
        \centering
        \includegraphics[width=\textwidth, height=0.9\textwidth]{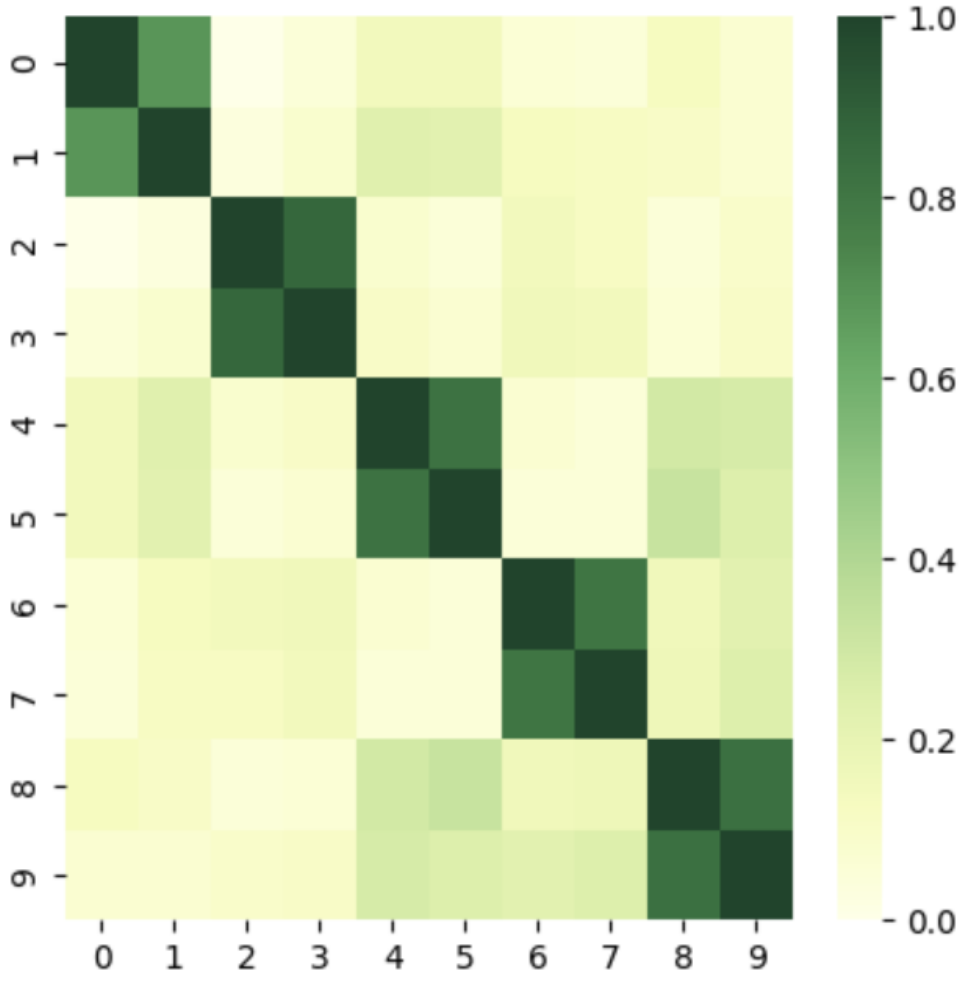} 
        \subcaption{Rank selection similarity}
        \vspace{0.25cm}
    \end{minipage}
    \begin{minipage}[b]{0.39\textwidth}
        \centering
        \includegraphics[width=0.98\textwidth, height=0.882\textwidth]{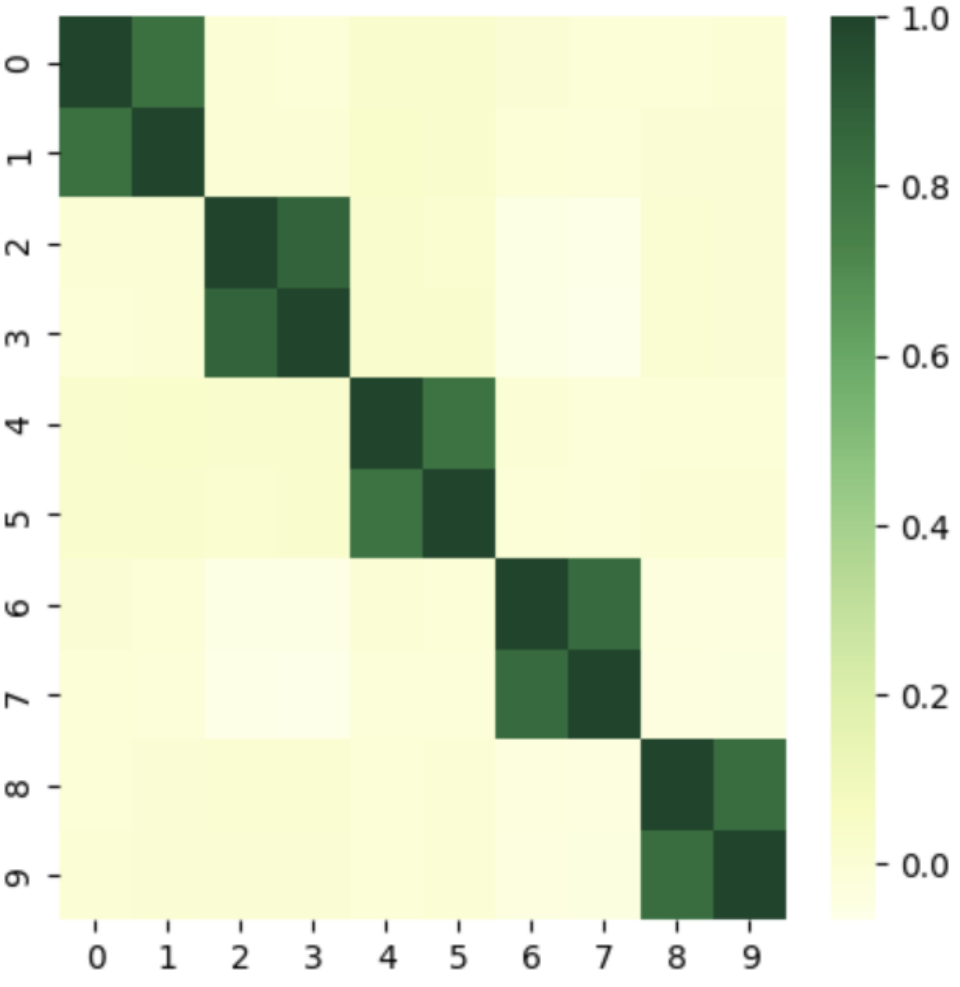} 
        \subcaption{Cosine similarity of local updates}
    \end{minipage}
    \caption{Visualization of similarity between clients. the x and y axes represent individual clients trained on 20 Newsgroups dataset with pathologic data distribution.}
    \label{fig:rank selection}
\end{figure}
\subsection{Analysis on Adaptive Rank Selection} \label{analysis_ars}

\begin{figure}[t]
    \centering
    \begin{minipage}[b]{0.23\textwidth}
        \centering
        \includegraphics[width=\textwidth]{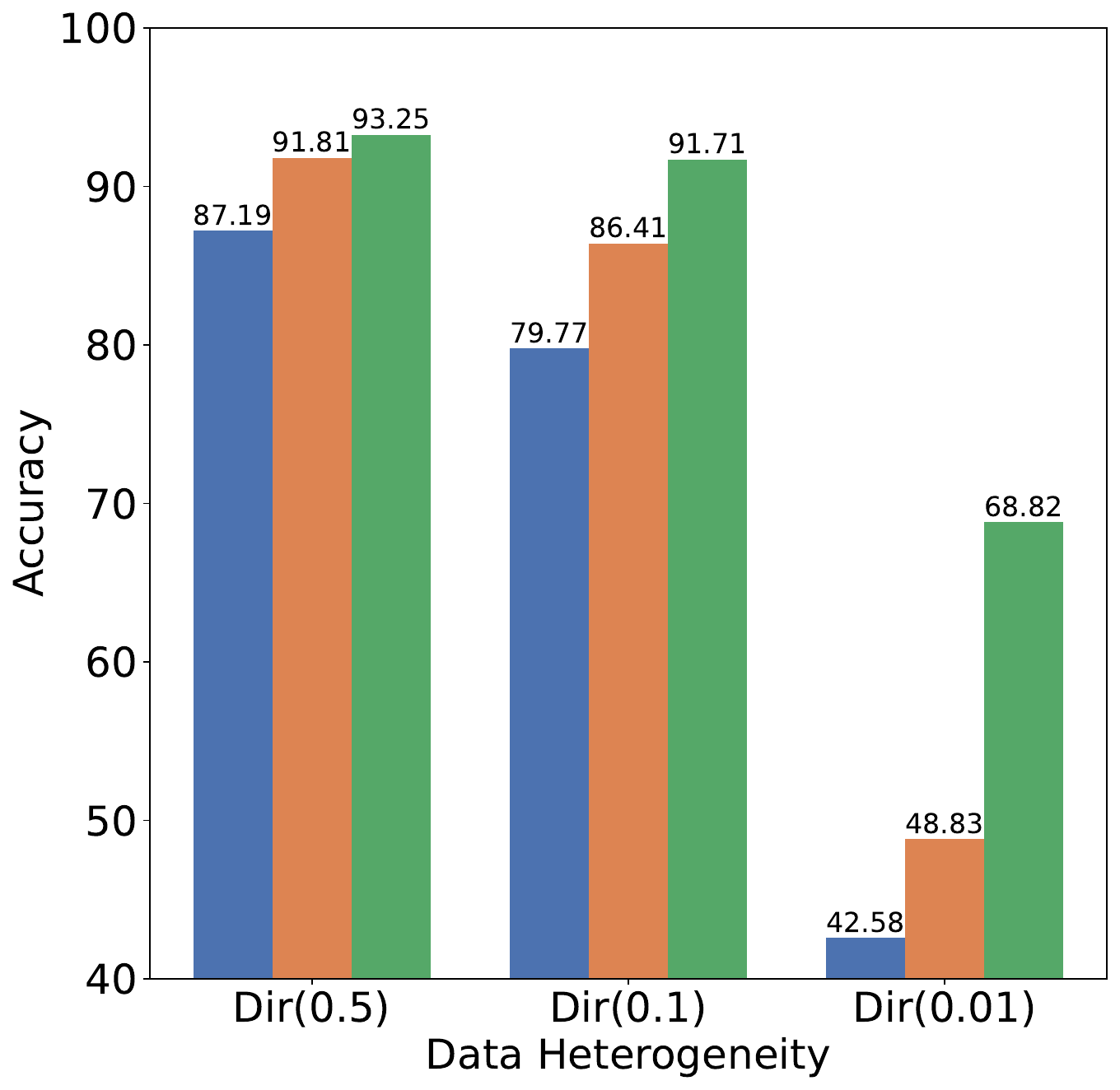}
        \subcaption{BANKING77}
    \end{minipage}
    \hfill
    \begin{minipage}[b]{0.23\textwidth}
        \centering
        \includegraphics[width=\textwidth]{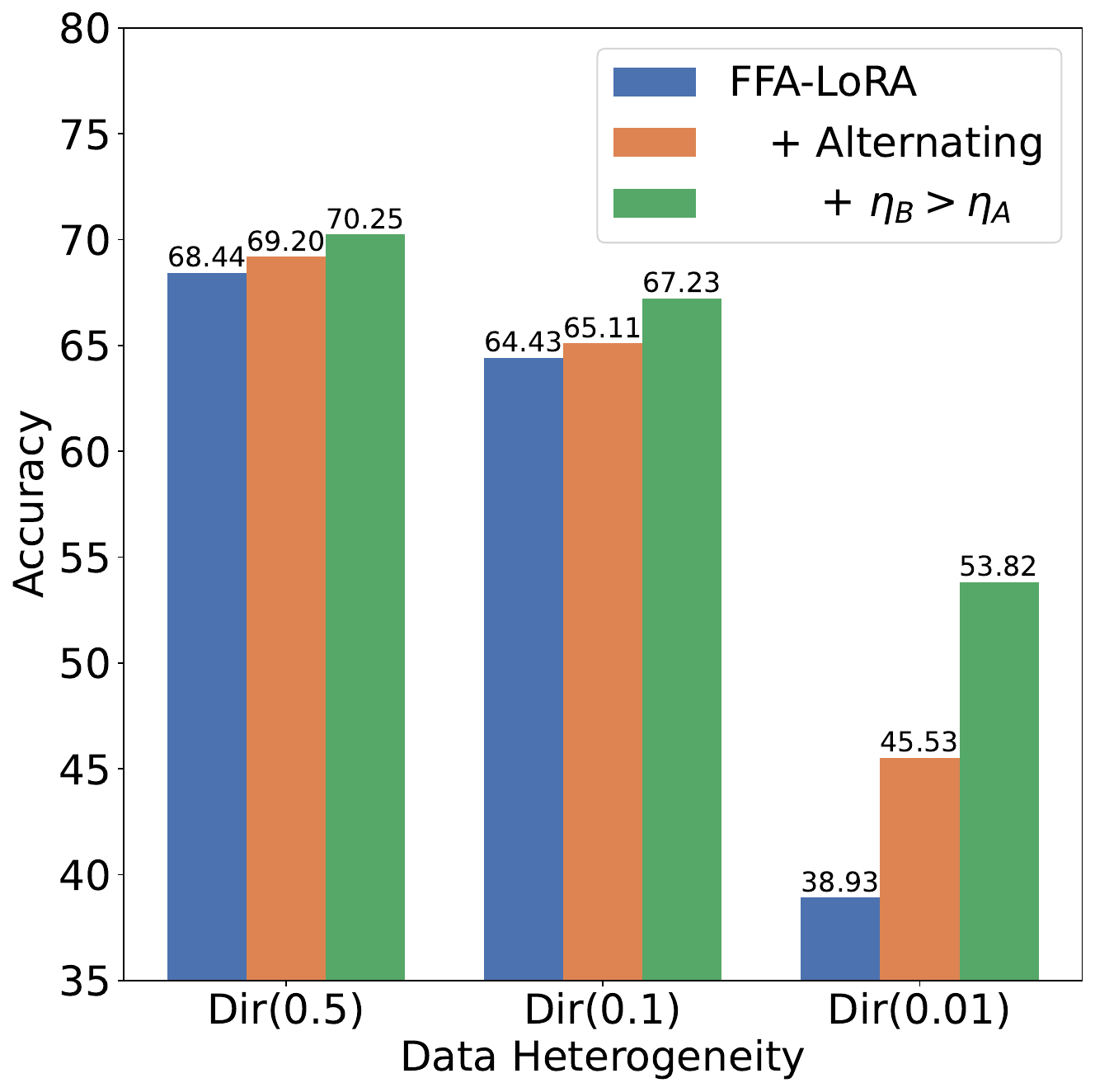}
        \subcaption{20 Newsgroups}
    \end{minipage}
    \caption{Ablation analysis for the effect of alternating freeze and learning rate adjustment under varying levels of heterogeneity.}
    \label{fig:ffa_alter}
\end{figure}

In this section,  we visualize the process of our adaptive rank selection, and explore how efficiently LoRA-A$^2$ trains and sends important ranks, highlighting the robustness of our algorithm in heterogeneous and low rank environments.
To simulate extreme cases of both identical and different client distributions, we test our algorithm on a pathological dataset using the 20 Newsgroups dataset. In this setup, 20 clients each holds data from only two classes, with consecutive pairs sharing the same classes, while others do not. For instance, clients 0 and 1 have classes "medical" and "space," whereas clients 2 and 3 have "motorcycle" and "religions". Detailed settings are shown in Appendix \ref{sec:appendix_c}.

\paragraph{Robustness to low-rank by adaptive module selection}
In this experiment, our algorithm selects $2\cdot N^{(m)}$ ranks from a total of $16 \cdot N^{(m)}$ across the whole RoBERTa model, guided by our importance criterion, and visualizes the adaptive selection of modules.
Figure \ref{fig:module selection} illustrates the number of ranks selected for each module in the model during the training. The figure shows that most modules are allocated zero ranks, indicating either no need for fine-tuning or the insignificance of updates on those modules. This suggests that our adaptive rank selection automatically prunes out modules that do not require additional fine-tuning.

To further justify that our adaptive rank selection successfully selects important modules, we conduct an ablation experiment on module selection, following the approach of AdaLoRA \cite{zhang2023adaptive} but in a federated setting. Figure \ref{fig:module ablation} displays the model's performance when only specific modules or layers are fine-tuned and other layers are frozen. The ablation experiment demonstrated that last layer in layer experiment and intermediate or output dense modules in module experiment led to the best performance, highlighting their importance for fine-tuning. This aligns with our findings, where the last layers and intermediate / output dense modules are automatically selected through adaptive rank selection, demonstrating the effectiveness of our algorithm in prioritizing essential modules for fine-tuning.

\paragraph{Robustness to data heterogeneity by client clustering}
Another effect of rank selection is the implicit clustering of clients to minimize conflicts among clients with dissimilar datasets and to enhance cooperation among those with similar ones.

Figure \ref{fig:rank selection} (a) illustrates how much local rank parameters are shared among different clients. The figure shows that clients with similar data distributions tend to share more rank parameters, while those with dissimilar data share fewer. This trend is also evident at the module level in Figure \ref{fig:module selection}, where clients 0 and 1 select a similar number of ranks for each module, differing from client 2, while retaining the tendency to choose more ranks from the last layers or intermediate and output dense modules.
These findings suggest that clients with similar datasets converge on the same ranks, facilitating cooperative training, whereas clients with dissimilar datasets select more distinct ranks, resulting in independent parameter updates.

Figure \ref{fig:rank selection} (b) further supports this by visualizing the cosine similarity between clients' model updates, which approaches 1 for clients with the same classes and remains close to 0 for those without data overlap. This underscores the cooperative nature of updates from similar clients while maintaining independence from dissimilar ones, thereby contributing to the robustness of our algorithm against data heterogeneity.

\subsection{Ablation Studies} \label{ablation}
The following ablation studies provide empirical evidence supporting our design choices for aggregation tactics and rank selection criteria.

\paragraph{Efficacy of alternating freeze}
To address the discordance problem in federated LoRA aggregation, we employ a strategy that alternately freezes LoRA modules $B$ and $A$, rather than freezing module $A$ only as in FFA-LoRA \citep{cho2023heterogeneous}. Furthermore, we set the learning rate of module $B$, $\eta_B$, to be five times that of module $A$, $\eta_A$, inspired by LoRA+ \cite{hayou2024loraplus}. This configuration further enhances overall performance and robustness, particularly in highly heterogeneous environments. Figure \ref{fig:ffa_alter} compares these approaches, showing that solely freezing $A$ is less effective under high data heterogeneity, whereas achieves consistently better performance.

\paragraph{Scalability and generalizability on model structures}
To evaluate the scalability and generalizability of our algorithm across various model structures, we present the experimental results on RoBERTa-large \citep{liu2019roberta} and DistilBERT \citep{sanh2019distilbert} models in Table \ref{tab:roberta-large} and Table \ref{tab:distilbert}, respectively. These tables illustrate the performance of our model when applied to diverse architectures and parameter configurations. The results show that our algorithm achieves superior performance, even on models with a larger number of parameters or different architectures. This highlights the robust scalability and generalizability of our approach across different model structures.

\begin{table}[t]
\centering
\begin{minipage}{0.48\textwidth}
\centering
\resizebox{\columnwidth}{!}{%
\begin{tabular}{c|ccc|c}
\toprule
\multirow{2}{*}{} \multirow{2}{*}{\begin{tabular}[c]{@{}c@{}}\# of\\ Ranks\end{tabular}} & \multicolumn{4}{c}{RoBERTa-Large}  \\
                 & FL+LoRA & FFA-LoRA & FlexLoRA$^*$ & Ours                                                                      \\ \midrule
8           & \underline{80.15}$_{\pm 0.58}$ & 62.98$_{\pm 0.61}$ & - & \textbf{85.98}$_{\pm 0.82}$ \\
4           & \underline{78.97}$_{\pm 0.52}$ & 62.45$_{\pm 0.33}$ & - & \textbf{84.62}$_{\pm 0.37}$ \\
2           & \underline{75.09}$_{\pm 1.20}$ & 61.55$_{\pm 1.05}$ & - & \textbf{83.40}$_{\pm 0.55}$ \\
1           & \underline{73.75}$_{\pm 1.53}$ & 58.06$_{\pm 1.90}$ & - & \textbf{85.66}$_{\pm 0.36}$ \\
\bottomrule
\end{tabular}%
}
\caption{Experimental results on RoBERTa-Large model. The level of heterogeneity is $Dir(0.01)$. \\
$^*$ FlexLoRA results could not be reported due to an ill-conditioned matrix issue in SVD decomposition.}
\label{tab:roberta-large}
\end{minipage}%
\vspace{0.3cm}
\hfill
\begin{minipage}{0.48\textwidth}
\centering
\resizebox{\columnwidth}{!}{%
\begin{tabular}{c|ccc|c}
\toprule
\multirow{2}{*}{} \multirow{2}{*}{\begin{tabular}[c]{@{}c@{}}\# of\\ Ranks\end{tabular}} & \multicolumn{4}{c}{DistilBERT}  \\
                 & FL+LoRA & FFA-LoRA & FlexLoRA & Ours                                                                      \\ \midrule
8           & 32.58$_{\pm 0.34}$ & 18.82$_{\pm 0.57}$ & \underline{51.21}$_{\pm 0.51}$ & \textbf{52.97}$_{\pm 0.32}$ \\
4           & 36.92$_{\pm 0.37}$ & 16.73$_{\pm 0.52}$ & \underline{41.26}$_{\pm 0.47}$ & \textbf{51.24}$_{\pm 0.44}$ \\
2           & 27.14$_{\pm 0.92}$ & 15.49$_{\pm 1.24}$ & \underline{34.05}$_{\pm 0.82}$ & \textbf{49.97}$_{\pm 0.33}$ \\
1           & \underline{21.59}$_{\pm 1.12}$ & 14.29$_{\pm 1.34}$ & 21.01$_{\pm 1.23}$ & \textbf{48.89}$_{\pm 0.41}$ \\
\bottomrule
\end{tabular}%
}
\caption{Experimental results on DistilBERT model. The level of heterogeneity is $Dir(0.01)$.}
\label{tab:distilbert}
\end{minipage}
\end{table}

\begin{table}[t]
\centering
\resizebox{\columnwidth}{!}{%
\begin{tabular}{c|ccc|c}
\toprule
$\epsilon$ & FL+LoRA & FFA-LoRA & FlexLoRA & Ours \\ \midrule
$\infty$    & 49.08$_{\pm 0.56}$ & 34.44$_{\pm 2.15}$ & \underline{55.24}$_{\pm 2.19}$ & \textbf{69.40}$_{\pm 0.48}$ \\
6    & 47.97$_{\pm 0.72}$ & 35.35$_{\pm 0.94}$ & \underline{50.22}$_{\pm 0.56}$ & \textbf{70.44}$_{\pm 1.88}$ \\
3    & 44.01$_{\pm 0.38}$ & 31.90$_{\pm 0.73}$ & \underline{49.62}$_{\pm 0.76}$ & \textbf{68.62}$_{\pm 1.61}$ \\
1    & 41.05$_{\pm 1.11}$ & 33.78$_{\pm 0.75}$ & \underline{49.39}$_{\pm 1.76}$ & \textbf{68.70}$_{\pm 0.22}$ \\
\bottomrule
\end{tabular}%
}
\caption{Experiments with differential privacy.}
    \label{tab:privacy}
\end{table}

\subsection{Additional Experiments}

\paragraph{Differential privacy}
According to \citet{sun2024improving}, discordance problem of federated LoRA intesnsified when Differential Privacy (DP) is applied \citep{10.1007/11681878_14, 10.1145/2976749.2978318}, due to the added noise amplifying errors. Specifically, if $\xi_B$ and $\xi_A$ stand for the noise added to $B$ and $A$, respectively, we have $\Delta W = (B + \xi_B) (A + \xi_A) = BA + B\xi_A + \xi_BA + \xi_B\xi_A$.

Table \ref{tab:privacy} represents experiments on BANKING77 dataset with DP. Following \citet{ryu2022appfl}, Laplace mechanism is adopted. The level of heterogeneity is $Dir(0.01)$ and the rank is set to 2 for each method. The clipping constant $C$ is set to either $2$ or $5$, whichever yields better performance, for each method. 

The tables demonstrates that FFA-LoRA \citep{sun2024improving}, FlexLoRA \citep{bai2024federated} and Our algorithm effectively mitigate the discordance problem, While FL with LoRA suffers from performance degradation. 
Moreover our algorithm shows the highest robustness under conditions of severe noise, such as $\epsilon = 1$ and $\epsilon = 3$, outperforming other baseline methods.

\paragraph{Computational overhead}
 Regarding computational overhead, our analysis shows that LoRA-A exhibits a 1.17x increase in computation time compared to standard FL+LoRA, slightly higher than FFA-LoRA (0.93x) and FlexLoRA (1x). This is due to gradient computation for local rank selection. However, we note that communication time, often the dominant bottleneck in federated learning, is significantly reduced by LoRA-A$^2$, outweighing the modest increase in computation time.

\paragraph{Other experiments}
 We also include further experiments addressing resource heterogeneity settings, pathological distributions, as well as investigations into convergence speed in Appendix \ref{sec:appendix_c}.

%% file: 6_conclusion.tex
In this work, we tackle the vulnerability of previous methods in high heterogeneity and low ranks by proposing a novel algorithm, LoRA-A$^2$, which shows robustness in these challenging conditions with alternating freeze and adaptive rank selection. Our approach offers significant improvements in communication efficiency without compromising performance, as demonstrated by a reduction of 99.8\% in parameter uploads compared to full fine-tuning. Through extensive experiments, we establish LoRA-A$^2$ as a superior alternative, providing a practical pathway for efficient and effective federated fine-tuning in diverse and resource-constrained environments.

%% file: acknowledgments.tex
This work was supported by Institute of Information \& communications Technology Planning \& Evaluation (IITP) grant funded by the Korea government(MSIT) (No.RS-2019-II191906, Artificial Intelligence Graduate School Program (POSTECH); RS-2021-II210739, Development of Distributed/Cooperative AI based 5G+ Network Data Analytics Functions and Control Technology; RS-2024-00457882, AI Research Hub Project; RS-2024-00509258, Global AI Frontier Lab).

%% file: 7_limitation.tex
LoRA-A$^2$ shows promising results and we plan to distribute the implementation code with detailed instructions for reproducibility. However, several areas remain open for future exploration.

First, our work mainly focuses on classification tasks, primarily due to computational constraints and the use of Dirichlet distribution to simulate non-IID conditions. However, extending LoRA-A$^2$ to more complex tasks, such as natural language generation, could offer additional perspectives. Future work with more resources could explore these broader applications.

Second, our experiments are primarily conducted on comparatively smaller language models, such as RoBERTa-base and RoBERTa-large, due to limited computation resources. Applying LoRA-A$^2$ to larger models, such as LLaMA or GPT-style architectures, could provide an opportunity to test its scalability. Investigating how well the method handles the increased parameter space of these state-of-the-art models could further demonstrate its efficiency.

Finally, due to the limited access to real world datasets, our current results are mainly based on simulated settings. Extensive research on real world dataset, which typically exhibit more diverse types of noise and heterogeneity would help understand performance and robustness of LoRA-A$^2$ in practical, dynamic environments.

%% file: appendix_a.tex
BANKING77 \citep{casanueva-etal-2020-efficient} is an intent classification dataset with 77 fine-grained intents related to the banking domain, comprising 10,003 training samples and 3,080 test samples. 20 Newsgroups \citep{Lang95} is a widely used text classification dataset with 20 classes, each representing a unique topic. It contains 11,314 training samples and 7,532 test samples.

We provide the statistics of two datasets in Table \ref{tab:news_stats} and Table \ref{tab:bank_stats}, respectively. $\mathcal{D}_k$ and $\left|\mathcal{C}_k\right|$ denotes the local dataset of $k$ and the number of unique classes in $\mathcal{D}_k$, respectively. Figure \ref{fig:dir_viz} shows the distribution of a local dataset for varying $\alpha$ simulating the Dirichlet distribution.

\begin{table}[t]
\centering
\resizebox{\columnwidth}{!}{%
\begin{tabular}{|c|cccccc|}
\hline
\multirow{2}{*}{}                    & \multicolumn{2}{c|}{$Dir(0.01)$}                       & \multicolumn{2}{c|}{$Dir(0.1)$}                        & \multicolumn{2}{c|}{$Dir(0.5)$}   \\ \cline{2-7} 
                                     & \multicolumn{1}{c|}{Train} & \multicolumn{1}{c|}{Test} & \multicolumn{1}{c|}{Train} & \multicolumn{1}{c|}{Test} & \multicolumn{1}{c|}{Train} & Test \\ \hline
$\max{\left|\left\{ \mathcal{D}_k \right\} \right|}_{k \in [K]}$ & \multicolumn{1}{c|}{1317}   & \multicolumn{1}{c|}{877}  & \multicolumn{1}{c|}{911}   & \multicolumn{1}{c|}{606}  & \multicolumn{1}{c|}{576}   & 383  \\ \hline
$\min{\left|\left\{ \mathcal{D}_k \right\} \right|}_{k \in [K]}$ & \multicolumn{1}{c|}{1}    & \multicolumn{1}{c|}{1}   & \multicolumn{1}{c|}{58}   & \multicolumn{1}{c|}{37}   & \multicolumn{1}{c|}{151}   & 100   \\ \hline
$\max{\left| \left\{ \mathcal{C}_k \right\} \right|}_{k \in [K]}$   & \multicolumn{1}{c|}{5}    & \multicolumn{1}{c|}{5}   & \multicolumn{1}{c|}{12}    & \multicolumn{1}{c|}{12}   & \multicolumn{1}{c|}{20}    & 14   \\ \hline
$\min{\left| \left\{ \mathcal{C}_k \right\} \right|}_{k \in [K]}$   & \multicolumn{1}{c|}{1}     & \multicolumn{1}{c|}{1}    & \multicolumn{1}{c|}{5}    & \multicolumn{1}{c|}{5}   & \multicolumn{1}{c|}{20}    & 12   \\ \hline
Number of classes                    & \multicolumn{6}{c|}{20}                                                                                                                             \\ \hline
Number of clients                    & \multicolumn{6}{c|}{30}                                                                                                                             \\ \hline
\end{tabular}
}
\caption{Statistics of 20 Newsgroups datasets.}
\label{tab:news_stats}
\end{table}

\begin{table}[t]
\centering
\resizebox{\columnwidth}{!}{%
\begin{tabular}{|c|cccccc|}
\hline
\multirow{2}{*}{}                    & \multicolumn{2}{c|}{$Dir(0.01)$}                       & \multicolumn{2}{c|}{$Dir(0.1)$}                        & \multicolumn{2}{c|}{$Dir(0.5)$}   \\ \cline{2-7} 
                                     & \multicolumn{1}{c|}{Train} & \multicolumn{1}{c|}{Test} & \multicolumn{1}{c|}{Train} & \multicolumn{1}{c|}{Test} & \multicolumn{1}{c|}{Train} & Test \\ \hline
$\max{\left|\left\{ \mathcal{D}_k \right\} \right|}_{k \in [K]}$ & \multicolumn{1}{c|}{639}   & \multicolumn{1}{c|}{212}  & \multicolumn{1}{c|}{672}   & \multicolumn{1}{c|}{185}  & \multicolumn{1}{c|}{473}   & 133  \\ \hline
$\min{\left|\left\{ \mathcal{D}_k \right\} \right|}_{k \in [K]}$ & \multicolumn{1}{c|}{50}    & \multicolumn{1}{c|}{30}   & \multicolumn{1}{c|}{139}   & \multicolumn{1}{c|}{43}   & \multicolumn{1}{c|}{248}   & 75   \\ \hline
$\max{\left| \left\{ \mathcal{C}_k \right\} \right|}_{k \in [K]}$   & \multicolumn{1}{c|}{15}    & \multicolumn{1}{c|}{10}   & \multicolumn{1}{c|}{34}    & \multicolumn{1}{c|}{24}   & \multicolumn{1}{c|}{65}    & 52   \\ \hline
$\min{\left| \left\{ \mathcal{C}_k \right\} \right|}_{k \in [K]}$   & \multicolumn{1}{c|}{2}     & \multicolumn{1}{c|}{2}    & \multicolumn{1}{c|}{18}    & \multicolumn{1}{c|}{15}   & \multicolumn{1}{c|}{37}    & 31   \\ \hline
Number of intents                    & \multicolumn{6}{c|}{77}                                                                                                                             \\ \hline
Number of clients                    & \multicolumn{6}{c|}{30}                                                                                                                             \\ \hline
\end{tabular}
}
\caption{Statistics of BANKING77 dataset.}
\label{tab:bank_stats}
\end{table}

%% file: appendix_b.tex
\paragraph{Hyperparameters} When training, we use AdamW \citep{loshchilov2018decoupled} optimizer with a learning rate of $\eta = 0.0005$. For LoRA-A$^2$, since $B$ and $A$ of each LoRA module are optimized separately, we use different learning rates for them. Specifically, $\eta_A = \eta$ is used for $A$ and $\eta_B = 5 \cdot \eta_A$ is used for $B$, which is inspired by LoRA+ \citep{hayou2024loraplus}. For HetLoRA, $\gamma = 0.99$ is used for the decaying factor as suggested by \citet{cho2023heterogeneous}. When evaluating, we merge the LoRA adapter $\Delta W$ with the pre-trained model $W_{0}$ using a scaling factor, so that $W_{ft} = W_{0} + \frac {16} {r} \Delta W$.

\paragraph{Implementation details} We simulate our FL setup using Flower \citep{beutel2020flower}, and utilize HuggingFace PEFT \citep{peft} library to train base models with LoRA. The base models are loaded using HuggingFace Transformers \citep{wolf-etal-2020-transformers} library. All experiments are conducted three times to ensure reproducibility, and the code will be released soon to promote transparency and support future research.

\begin{figure}[t!]
    \centering
    \begin{minipage}[b]{0.23\textwidth}
        \centering        \includegraphics[width=\textwidth]{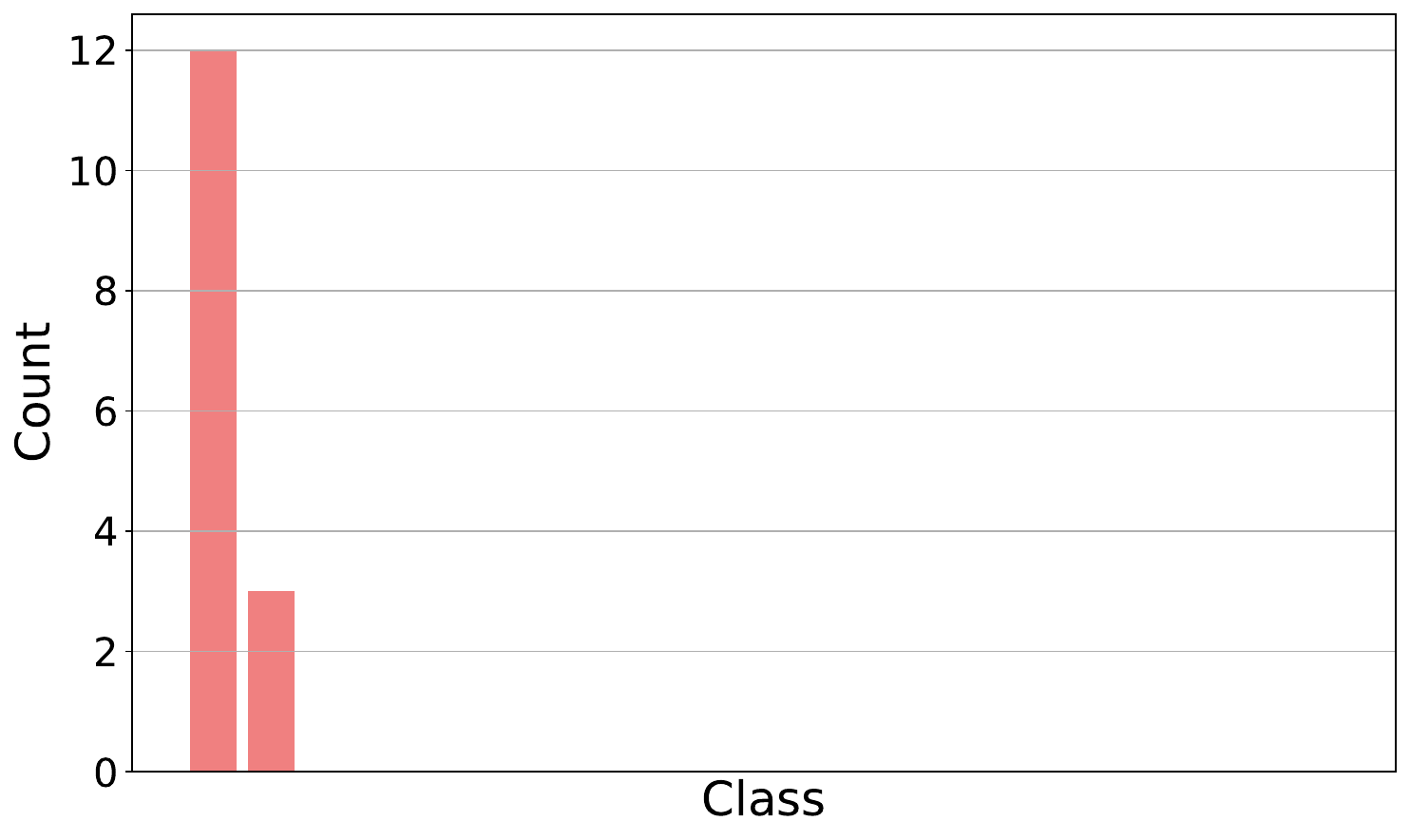}
        \subcaption{$\alpha = 0.01$}
    \end{minipage}
    \hfill
    \begin{minipage}[b]{0.23\textwidth}
        \centering
        \includegraphics[width=\textwidth]{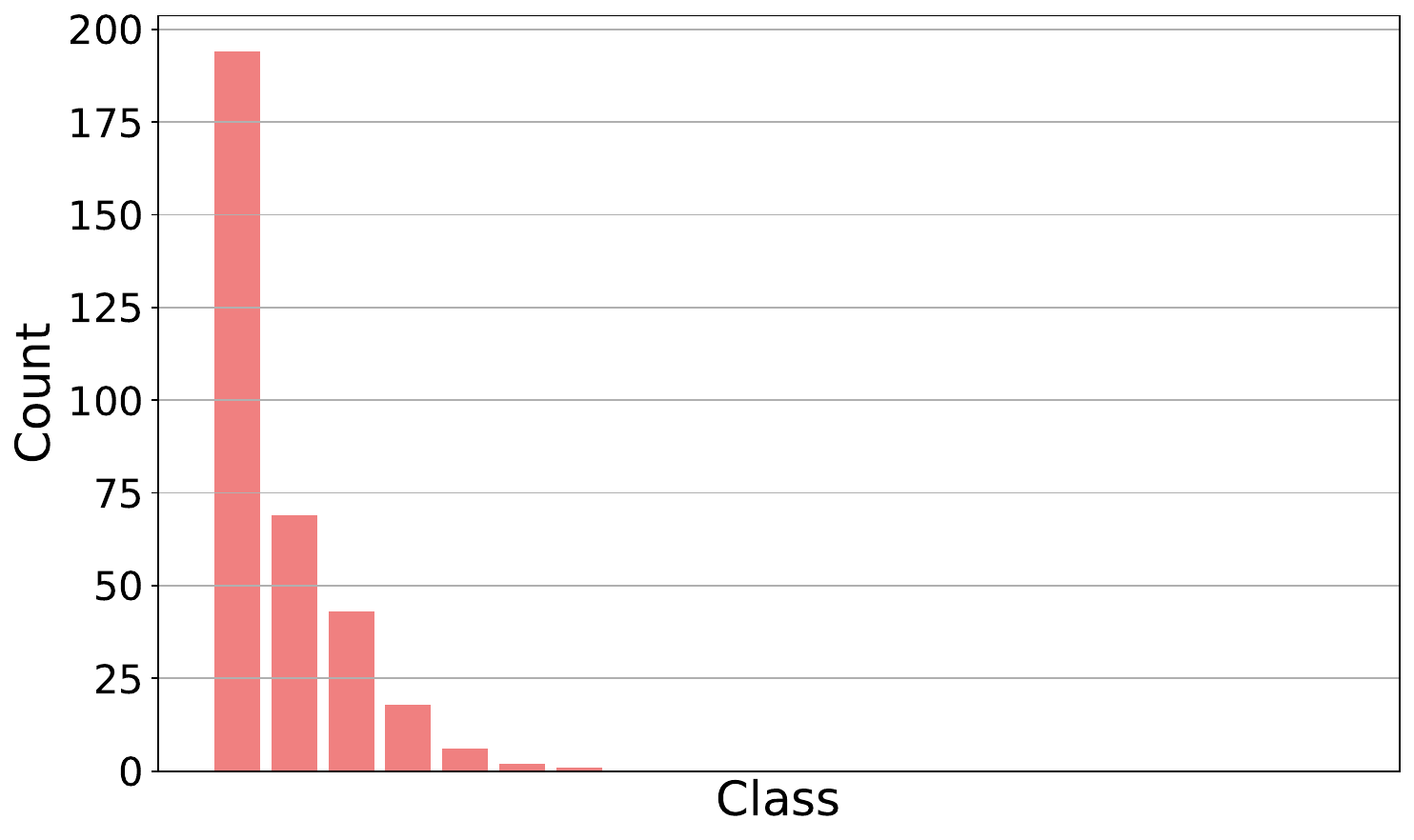}
        \subcaption{$\alpha = 0.1$}
    \end{minipage}
    \hfill
    \begin{minipage}[b]{0.23\textwidth}
        \centering
        \includegraphics[width=\textwidth]{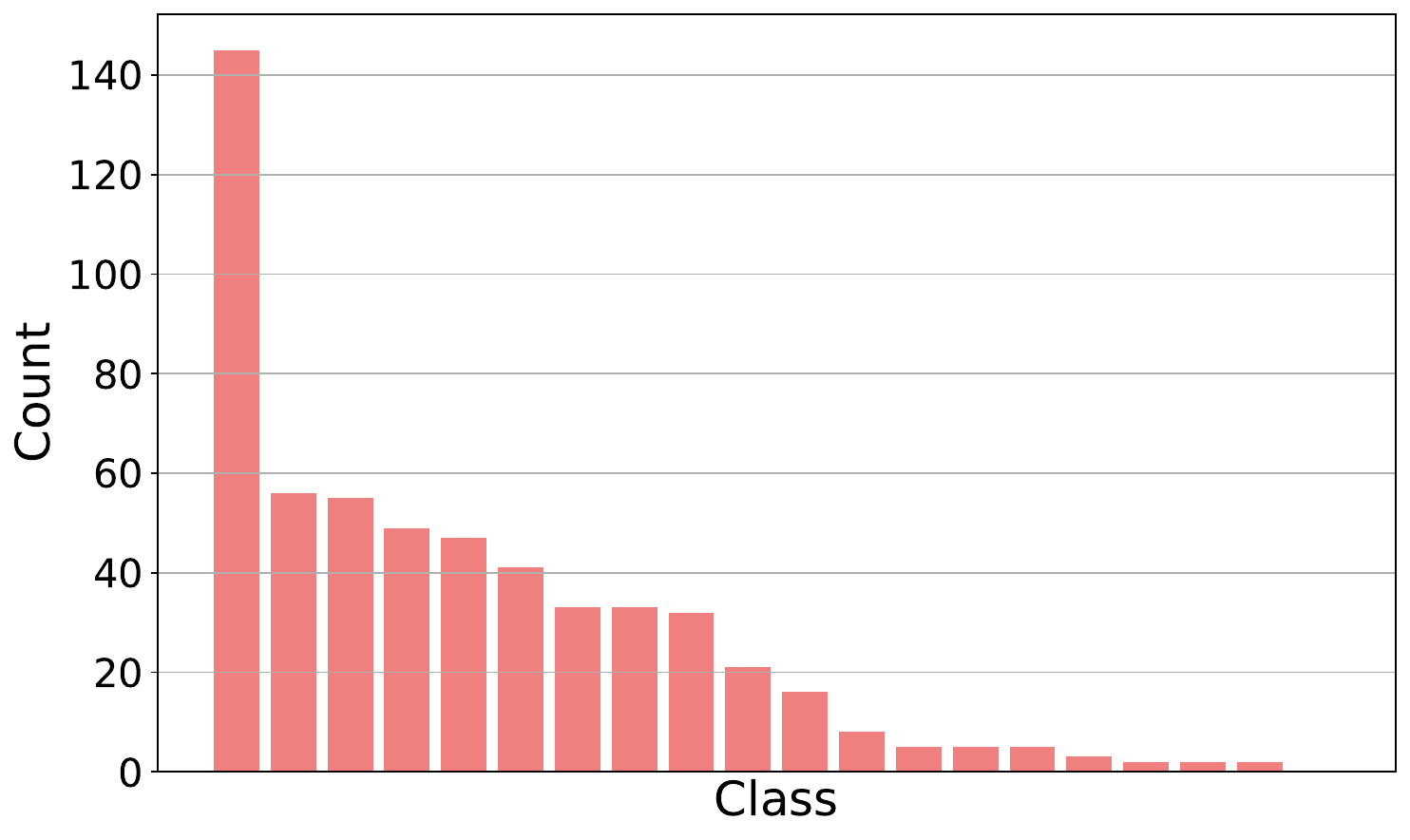}
        \subcaption{$\alpha = 0.5$}
    \end{minipage}
    \hfill
    \begin{minipage}[b]{0.23\textwidth}
        \centering
        \includegraphics[width=\textwidth]{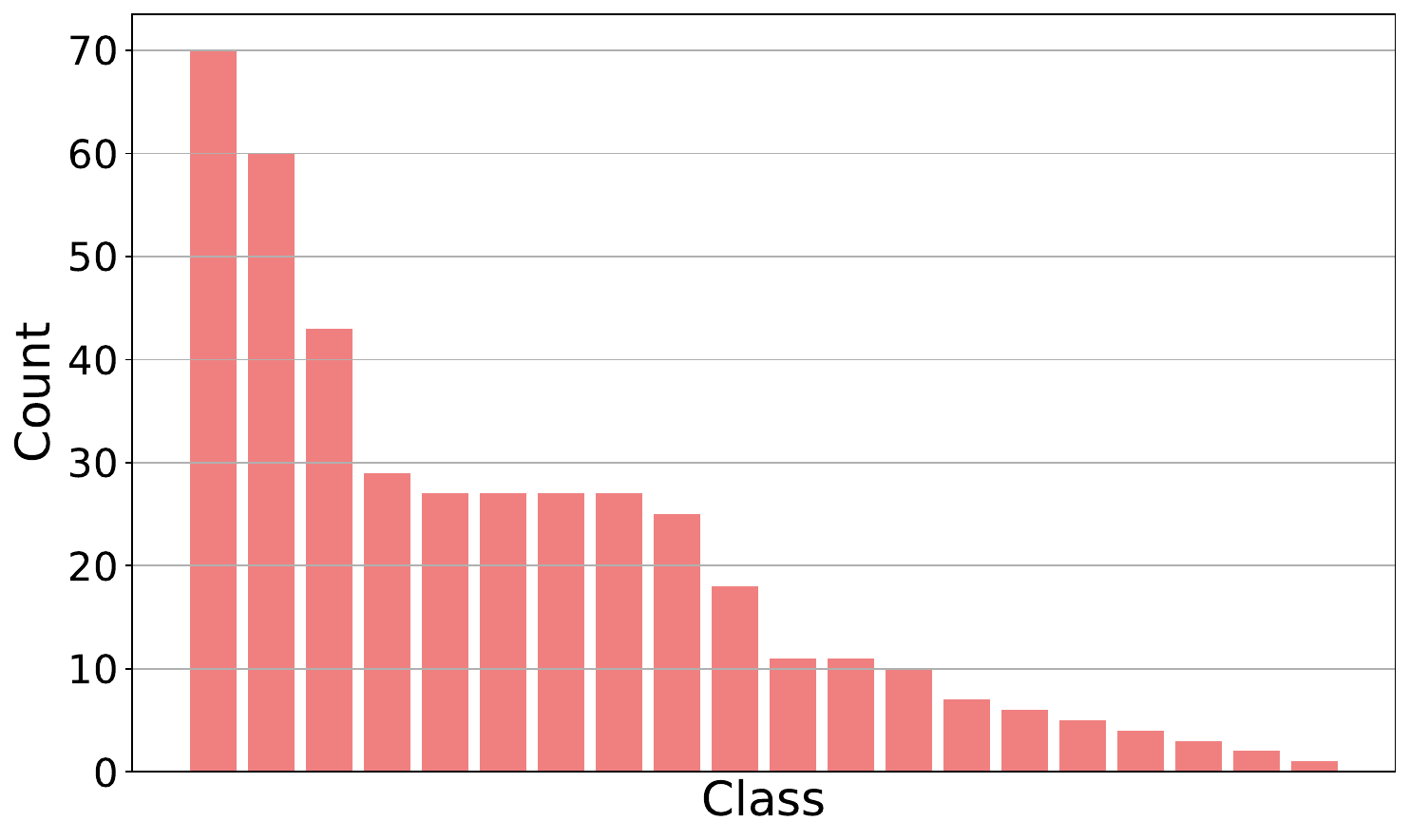}
        \subcaption{$\alpha = 1$}
    \end{minipage}
    \caption{Local distribution of client 0 for different $Dir(\alpha)$ on 20 Newsgroup dataset experiments.}
    \label{fig:dir_viz}
\end{figure}

\begin{table}[t]
\centering
\resizebox{\columnwidth}{!}{%
\begin{tabular}{c|ccc|c}
\toprule
Rank & FL+LoRA & FFA-LoRA & FlexLoRA & Ours \\ \midrule
8    & 53.80$_{\pm 1.44}$ & 52.60$_{\pm 0.96}$ & \textbf{60.36}$_{\pm 1.15}$ & \underline{58.74}$_{\pm 0.95}$ \\
4    & 55.03$_{\pm 0.43}$ & 50.57$_{\pm 1.58}$ & \textbf{59.12}$_{\pm 0.98}$ & \underline{58.62}$_{\pm 1.51}$ \\
2    & 50.40$_{\pm 0.77}$ & 48.36$_{\pm 0.86}$ & \underline{55.46}$_{\pm 0.99}$ & \textbf{59.63}$_{\pm 0.59}$ \\
1    & \underline{51.24}$_{\pm 3.12}$ & 46.92$_{\pm 1.30}$ & 51.05$_{\pm 0.69}$ & \textbf{59.11}$_{\pm 0.88}$ \\
\bottomrule
\end{tabular}%
}
\caption{Experiments on pathologic settings.}
    \label{tab:pathologic}
\end{table}

%% file: appendix_c.tex
\begin{figure*}[htbp]
    \centering
    \begin{minipage}[b]{0.3\textwidth}
        \centering
        \includegraphics[width=\textwidth]{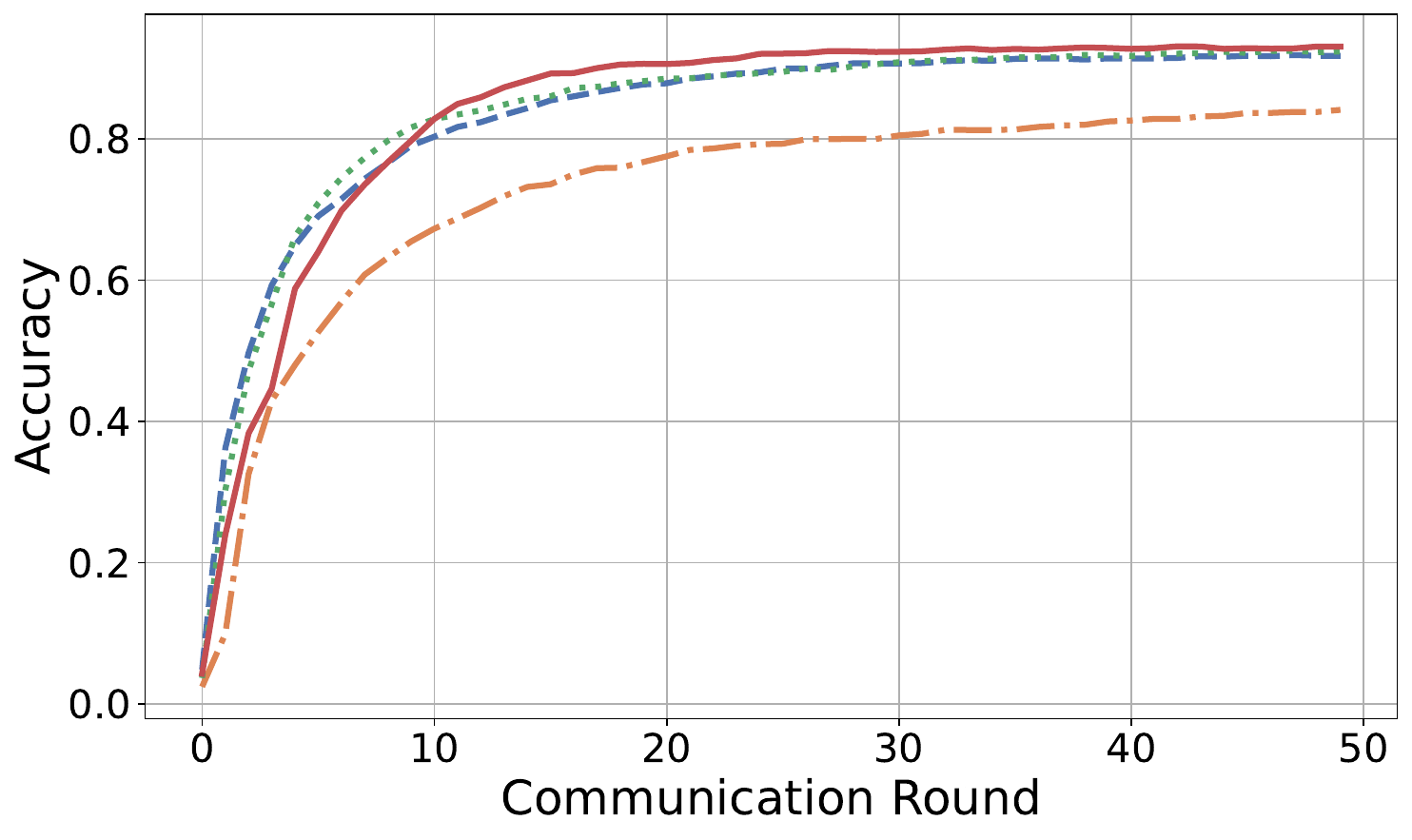}
        \subcaption{$Dir(0.5)$}
    \end{minipage}
    \hfill
    \begin{minipage}[b]{0.3\textwidth}
        \centering
        \includegraphics[width=\textwidth]{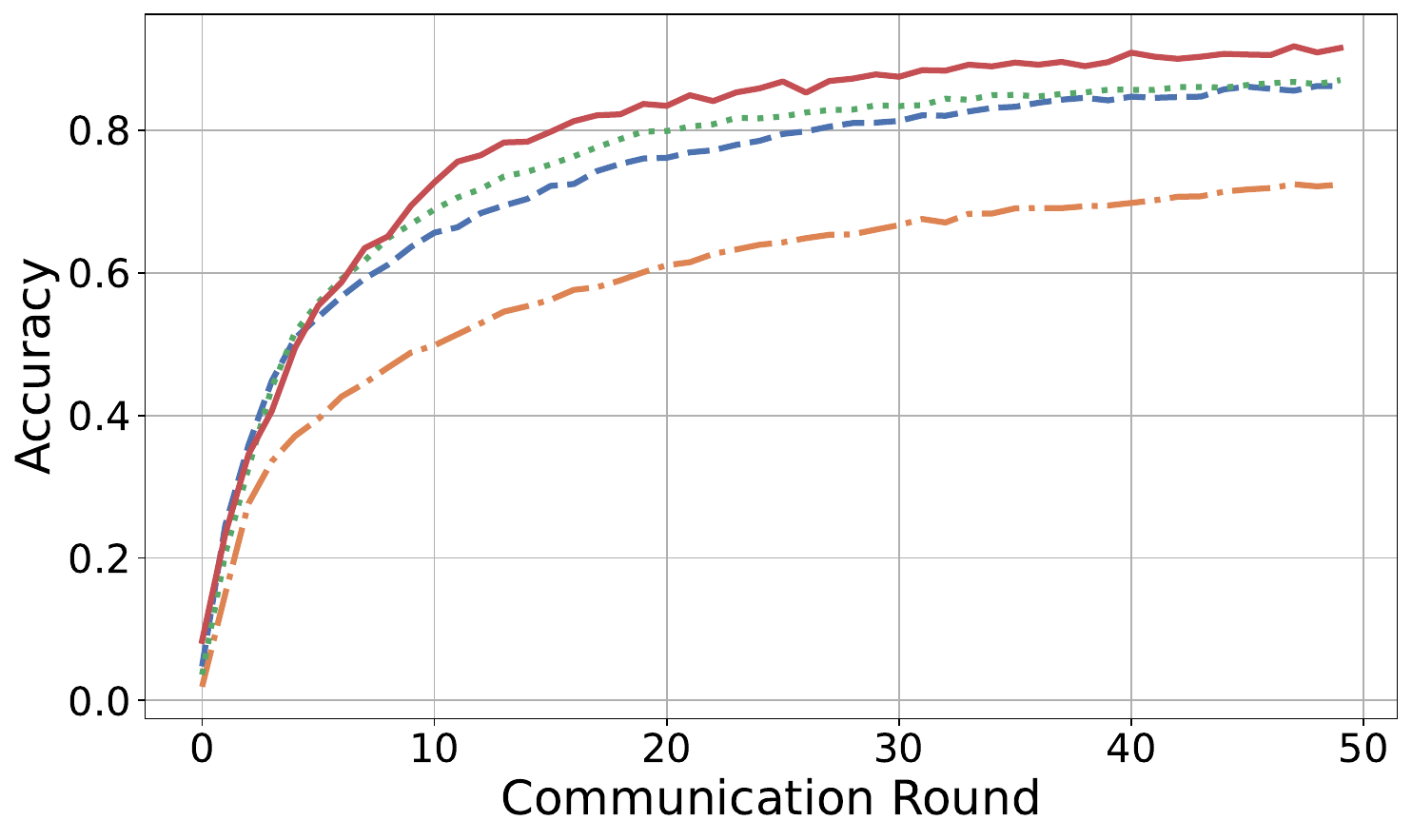}
        \subcaption{$Dir(0.1)$}
    \end{minipage}
    \hfill
    \begin{minipage}[b]{0.3\textwidth}
        \centering
        \includegraphics[width=\textwidth]{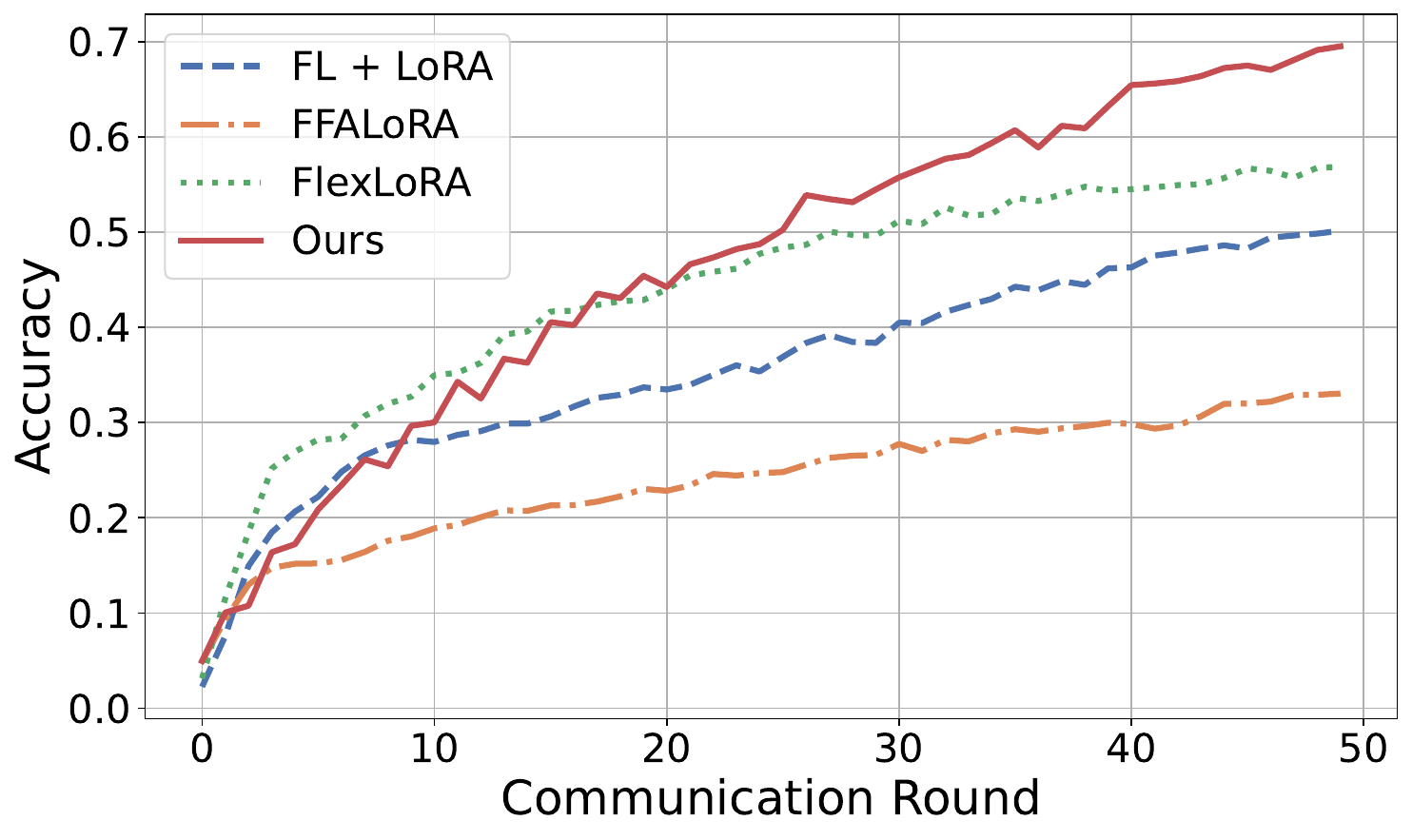}
        \subcaption{$Dir(0.01)$}
    \end{minipage}
    \caption{Convergence curve of baseline methods in various levels of heterogeneity. Experimented on BANKING77 dataset, with local ranks all set to 2.}
    \label{fig:convergence}
\end{figure*}

\paragraph{Pathologic setting} Table \ref{tab:pathologic} provides experiments on pathologic setting, which is also used to generate Figure \ref{fig:rank selection} in Section \ref{analysis_ars}, to show the efficacy of adaptive rank selection. In this setting, we have $K = 20$ clients. And client $(2k - 1)$ and client $(2k)$ exclusively possess half of class $(2k - 1)$ and $(2k)$ of 20 Newsgroups datasets, respectively, for $k = 1, 2, \cdots, 10$.

\paragraph{Convergence speed analysis}
Figure \ref{fig:convergence} shows the convergence curves of our algorithm and baselines. The figure demonstrates that our algorithm shows similar convergence speed compared to baseline methods in various levels of heterogeneity.

\begin{figure}[t]
    \centering
    \includegraphics[width=0.45\textwidth]{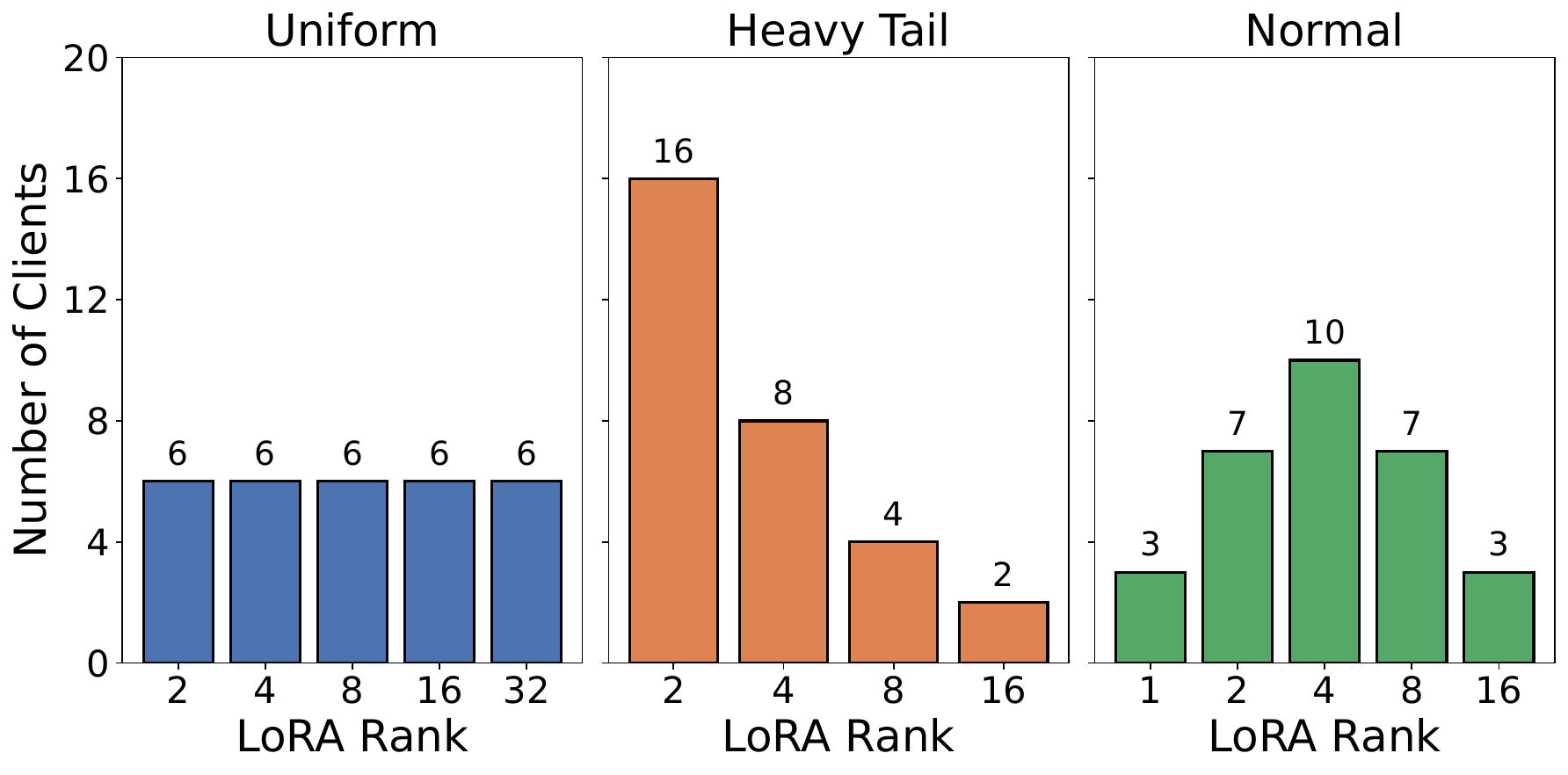} 
    \caption{Three types of simulated client rank distributions used to evaluate performance under resource heterogeneity settings.}
    \label{fig:het_dist}
\end{figure}

\paragraph{Resource heterogeneity} \label{sec:erh}
In this experiment, we analyze the efficacy of our algorithm under varying resource constraints across clients. Specifically, we assume that each client has a different communication cost budget \citep{pmlr-v202-chen23aj} and has different maximum LoRA rank for its adapter. Following \citet{bai2024federated}, we simulate three types of resource heterogeneity, as illustrated in Figure \ref{fig:het_dist}.

In Table \ref{tab:resource_table}, we compare our method against HetLoRA and FlexLoRA, two previous LoRA methods that can handle resource heterogeneity in FL. The experimental results demonstrates that our algorithm shows slightly better or similar performance compared to HetLoRA with less number of communicated parameters.

\begin{table}[t]
\centering
\resizebox{\columnwidth}{!}{%
\begin{tabular}{ccccc}
\toprule
\multirow{2}{*}{Distribution} & \multirow{2}{*}{Method} & \multicolumn{2}{c}{BANKING77 Dataset} & \multirow{2}{*}{\begin{tabular}[c]{@{}c@{}}Communicated\\ Parameters\end{tabular}} \\
& & $Dir(0.1)$ & $Dir(0.01)$ & \\ \midrule
\multirow{3}{*}{Uniform} 
& HetLoRA & \underline{86.91}$_{\pm 0.43}$ & \underline{68.53}$_{\pm 2.14}$ & 3.09B \\  
& FlexLoRA & 73.01$_{\pm 0.69}$ & 45.41$_{\pm 1.60}$ & 3.09B \\ 
\cmidrule{2-5}
& Ours & \textbf{92.02}$_{\pm 0.16}$ & \textbf{70.67}$_{\pm 0.76}$ & 1.97B \\ \midrule
\multirow{3}{*}{Heavy Tail} 
& HetLoRA & \underline{85.82}$_{\pm 0.54}$ & \underline{69.57}$_{\pm 1.13}$ & 1.06B \\  
& FlexLoRA & 82.69$_{\pm 0.86}$ & 52.46$_{\pm 1.72}$ & 1.06B \\ 
\cmidrule{2-5}
& Ours & \textbf{91.72}$_{\pm 0.07}$ & \textbf{69.95}$_{\pm 2.23}$ & 0.942B \\  \midrule
\multirow{3}{*}{Normal} 
& HetLoRA & 84.57$_{\pm 0.55}$ & \textbf{70.34}$_{\pm 0.15}$ & 1.34B \\  
& FlexLoRA & 77.08$_{\pm 0.68}$ & 53.37$_{\pm 3.49}$ & 1.34B \\ 
\cmidrule{2-5}
& Ours & \textbf{92.08}$_{\pm 0.18}$ & \underline{69.04}$_{\pm 0.64}$ & 0.932B \\ 
\bottomrule
\end{tabular}%
}
\caption{Experimental results under resource heterogeneity settings.}
\label{tab:resource_table}
\end{table}

\paragraph{Efficacy of importance criterion}
As mentioned in Section \ref{ars}, other criteria such as magnitude-based or importance-based scoring functions can be used for selecting ranks. Table \ref{tab:criteria_table} shows that our criterion outperforms others, with less communication than the magnitude-based criterion.

\begin{table}[t]
\centering
\resizebox{\columnwidth}{!}{%
\begin{tabular}{cccc}
\toprule
\multirow{2}{*}{} & \multicolumn{2}{c}{BANKING77 Dataset} & \multirow{2}{*}{\begin{tabular}[c]{@{}c@{}}Communicated\\ Parameters\end{tabular}} \\
                  & $Dir(0.1)$            & $Dir(0.01)$          &                                                                       \\ \midrule
Importance           & 91.29$_{\pm 0.76}$                       & 66.92$_{\pm 1.58}$   & 0.215B                                                                \\ 
Magnitude          & \underline{91.71}$_{\pm 0.23}$                         & \underline{68.00}$_{\pm 0.57}$   & 0.651B                                                                \\ \midrule
Ours              & \textbf{92.02}$_{\pm 0.36}$                       & \textbf{69.40}$_{\pm 0.48}$   & 0.507B                                                                                                                       \\ \bottomrule
\end{tabular}%
}
\caption{Ablation study on scoring functions.}
\label{tab:criteria_table}
\end{table}

\begin{figure}[t]
    \centering
    \includegraphics[width=0.4\textwidth, height=0.35\textwidth]{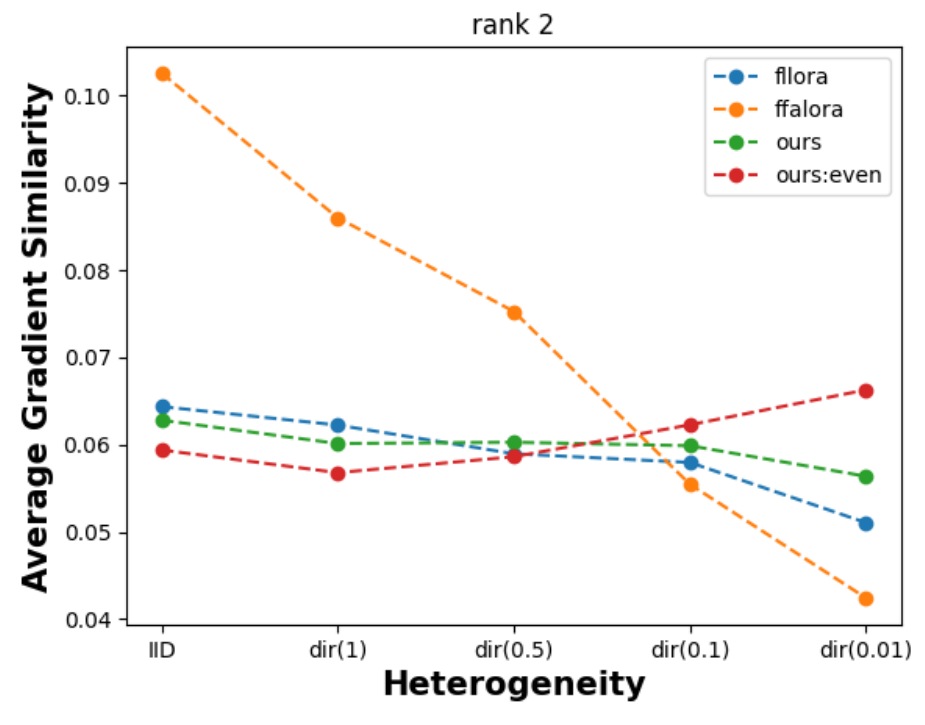} 
    \caption{Average Gradient Similarity on various level of heterogeneity. Experimented on 20 Newsgroups dataset and the ranks were all set to 2.}
    \label{fig:ags}
\end{figure}

\paragraph{Client drift experiment} \label{sec:CDE}
To thoroughly analyze the impact of data heterogeneity within constrained parameter spaces, we conducted additional experiments that illustrate the local client drift observed in baseline methods operating under these limitations. We quantified the degree of client drift by calculating the "Average Gradient Similarity," defined as follows:
\begin{equation}
\begin{aligned}
    &Average Gradient Similarity = \\
    &\frac{1}{n^2}\sum_i^n \sum_i^n \frac{(\Delta W^t_i - \Delta W^{t-1}_i) \cdot (\Delta W^t_j - \Delta W^{t-1}_j)}{||\Delta W^t_i - \Delta W^{t-1}_i ||\cdot||\Delta W^t_j - \Delta W^{t-1}_j ||}
\end{aligned}
\label{eq:AGS}
\end{equation}

The experimental results presented in Figure \ref{fig:ags} indicate a rapid decline in average gradient similarity as the level of heterogeneity increases. In contrast, our method demonstrates greater robustness, exhibiting lower client drift even in rounds where only the LoRA module A is updated. These findings are consistent with the results shown in Figure \ref{fig:volnerability} and Table \ref{tab:merged_table}, which illustrate that FFA-LoRA experiences the most significant performance decline between the directional settings of 0.1 and 0.01, while our algorithm maintains its effectiveness in heterogeneous environments.

%% file: appendix_d.tex
Here's brief proof for the proposition made in section \ref{subsec:theory}:

Proof) First, since FFA-LoRA freezes all the $A_i$'s permanently, $\Omega_\text{FFA-LoRA} = \left\{ B_i \right\}_{i=1}^N$. Next, since FL + LoRA and FlexLoRA update $B_i$'s and $A_i$'s simultaneously, $\Omega_\text{FL + LoRA} = \left\{ \left(B_i, A_i\right) \right\}_{i=1}^N = \Omega_\text{FlexLoRA}$. Finally, $\Omega_{\text{LoRA-A}^2}= \left\{ \left(\bar{B}_i, \bar{A}_i\right) \right\}_{i=1}^N$, where its subspace $\left\{B_i \right\}_{i=1}^N$ or $\left\{ A_i \right\}_{i=1}^N$ is optimized according to the Alternating freeze and Adaptive rank selection algorithm.  Therefore, noting that $r \leq r_G$, we have $\Omega_\text{FFA-LoRA} \subsetneq \Omega_\text{FL + LoRA} = \Omega_\text{FlexLoRA} \subset \Omega_{\text{LoRA-A}^2}$. $\square$